\documentclass[lettersize,journal]{IEEEtran}
\usepackage{amsmath,amsfonts}

\usepackage{algorithm}
\usepackage{array}
\usepackage[caption=false,font=normalsize,labelfont=sf,textfont=sf]{subfig}
\usepackage{textcomp}
\usepackage{stfloats}
\usepackage{url}
\usepackage{verbatim}
\usepackage{graphicx}
\usepackage{cite}
\usepackage{xcolor}
\usepackage{tabularx}
\usepackage{booktabs}
\usepackage{multirow,makecell,array}
\usepackage{threeparttable}
\usepackage{algpseudocode}
\usepackage{mathtools}
\usepackage{float}
\usepackage{csquotes}
\usepackage{colortbl}
\usepackage{ulem}
\usepackage{orcidlink}
\usepackage{soul}
\usepackage{makecell}
\usepackage{arydshln}
\soulregister\ref7

\begin{document}

\title{Towards Head Computed Tomography Image Reconstruction Standardization with Deep Learning Assisted Automatic Detection}

\author{Bowen Zheng~\orcidlink{0009-0000-3167-4399}, Chenxi Huang~\orcidlink{0000-0003-2863-6089}, Yuemei Luo~\orcidlink{0000-0001-8741-449X}
}

\markboth{Journal of \LaTeX\ Class Files,~Vol.~14, No.~8, August~2021}%
{Shell \MakeLowercase{\textit{et al.}}: A Sample Article Using IEEEtran.cls for IEEE Journals}


\maketitle

\begin{abstract}
Three-dimensional (3D) reconstruction of head Computed Tomography (CT) images elucidates the intricate spatial relationships of tissue structures, thereby assisting in accurate diagnosis. Nonetheless, securing an optimal head CT scan without deviation is challenging in clinical settings, owing to poor positioning by technicians, patient's physical constraints, or CT scanner tilt angle restrictions. Manual formatting and reconstruction not only introduce subjectivity but also strain time and labor resources. To address these issues, we propose an efficient automatic head CT images 3D reconstruction method, improving accuracy and repeatability, as well as diminishing manual intervention. Our approach employs a deep learning-based object detection algorithm, identifying and evaluating orbitomeatal line landmarks to automatically reformat the images prior to reconstruction. Given the dearth of existing evaluations of object detection algorithms in the context of head CT images, we compared 12 methods from both theoretical and experimental perspectives. By exploring their precision, efficiency, and robustness, we singled out the lightweight YOLOv8 as the aptest algorithm for our task, with an mAP of 92.77\% and impressive robustness against class imbalance. Our qualitative evaluation of standardized reconstruction results demonstrates the clinical practicability and validity of our method.
\end{abstract}


\section{Introduction}
Three-dimensional (3D) reconstruction of computed tomography images has become an indispensable instrument in a broad spectrum of clinical tasks, particularly over the last few years\cite{bb1}\cite{bb3}\cite{bb4}. Specifically, head CT scans hold a pivotal role, revealing key details about structures such as the brain, skull, sinuses, and various soft tissues. This provides crucial insights instrumental in diagnosing and treatments for a wide range of conditions, from traumas and tumors to strokes and sinusitis\cite{bb5}\cite{bb6}. However, manual interpretation of these scans often proves time-consuming and error-prone. Historically, this process has been fraught with potential pitfalls due to human error and variability in diagnostic interpretation. This is primarily attributed to the intricate details and subtle variations that can signify different pathological conditions\cite{bb7}\cite{bb8}. 

Employing computer 3D reconstruction software is possible to transform a sequence of CT axial images into a comprehensive three-dimensional model. This process transforms the series of two-dimensional scans into an intuitive representation of the spatial relationships among tissue structures. It's a transformative leap from the conventional methods but isn't devoid of challenges. The outcome facilitates precise surgical planning and patient-specific treatment strategies, a significant enhancement in medical procedures. The current quality control standards for head CT images take the orbitomeatal line as the baseline for scans. This line is drawn from the center of the external auditory canal (EAC) to the outer canthus of the ipsilateral eye\cite{bb9}.

Nonetheless, it is often challenging to achieve ideal head CT images free of lateral deviations and based on the orbitomeatal line. Traditional solutions have been more reactive than proactive in this respect. Factors such as poor positioning by technicians, patient's physical constraints, and restrictions in the tilt angle of the CT scanner contribute to this challenge. As a result, radiologic technologists frequently manually reformat axial images using thin-layer data post-CT examination. This makeshift solution, while practical, isn't optimal. Unfortunately, this process may precipitate deviations in the reconstruction baseline as a consequence of the operator’s subjective judgment, thereby affecting the accuracy of reconstruction outcomes. Additionally, manual reformatting and the subsequent 3D reconstruction require a significant investment in terms of time and labor.

Deep learning has emerged as a vital player in fields such as object detection, medical image segmentation\cite{bb10}\cite{bb11}\cite{bb34}, diagnostics\cite{bb12}\cite{bb13}, and predictive analytics\cite{bb14}\cite{bb15}\cite{bb35}. Yet, its full potential, especially in the domain of CT image reconstruction, remains largely untapped. Previous research proposed a semi-automatic multiplanar reconstruction method\cite{bb16}. This method mandates manually setting five head landmarks on axial images to identify the orbitomeatal line, facilitating 3D and multiplanar reconstruction. However, it heightens the labor intensity given the necessity of manual landmark identification. 

Another approach utilizes an object detection algorithm to automatically reformat head CT images based on the orbitomeatal line\cite{bb17}. Despite its benefits, this method mainly focuses on the automatic reformatting of axial head CT images, with limited involvement in 3D reconstruction. Moreover, the You Only Look Once (YOLO) model used in this object detection algorithm exhibits an accuracy of only 0.68, suggesting substantial potential for improvement in detection accuracy.

Navigating these prevailing gaps, we introduce a robust and trustworthy automated 3D reconstruction method, a solution that could substantially enhance diagnostic precision. The primary contributions of this paper are as follows.

\begin{enumerate}
    \item We have proposed an automatic 3D reconstruction method for head CT images, utilizing automatically reformatted CT images. This method not only promises more accurate and repeatable results, mitigating potential inconsistencies arising from varying technicians' manual reconstruction judgments, but also curtails the need for manual input, yielding considerable savings in time and labor.
    \item Capitalizing on a deep learning-based object detection algorithm, we have devised a streamlined process for 3D reconstruction. This algorithm identifies and evaluates orbitomeatal line landmarks, typically manually annotated by radiologists, to automate image reformatting. This process significantly reduces labor expenditure associated with image reformatting.
    \item We conducted an exhaustive comparative analysis of 12 deep learning-based object detection methods, focusing on their accuracy, efficiency, and robustness in the context of automatic reformatting of CT images. Grounded in both theoretical and experimental insights, we discern the most effective and efficient algorithm for this particular task.
\end{enumerate}

\begin{figure*}[htbp]
  \centering
  \includegraphics[width=\textwidth]{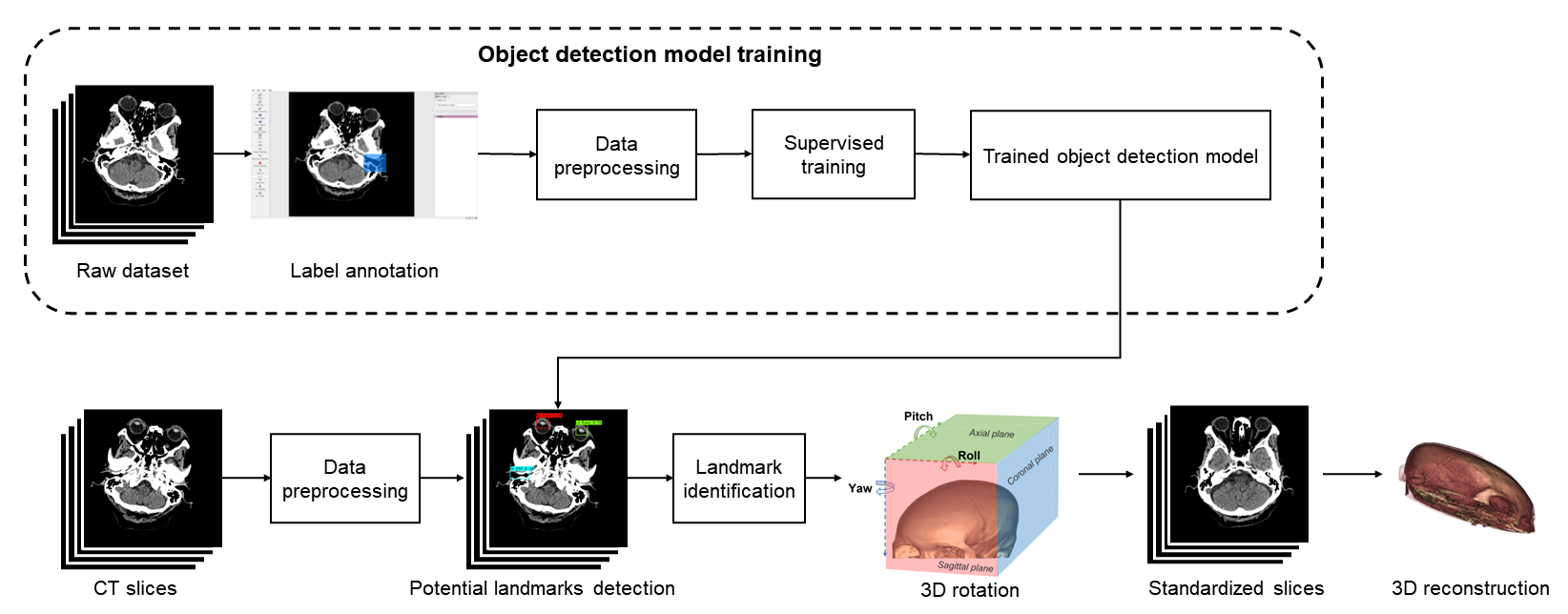}
  \caption{Proposed approach}
  \label{f1}
\end{figure*}

\begin{figure*}[htbp]
  \centering
  \includegraphics[width=\textwidth]{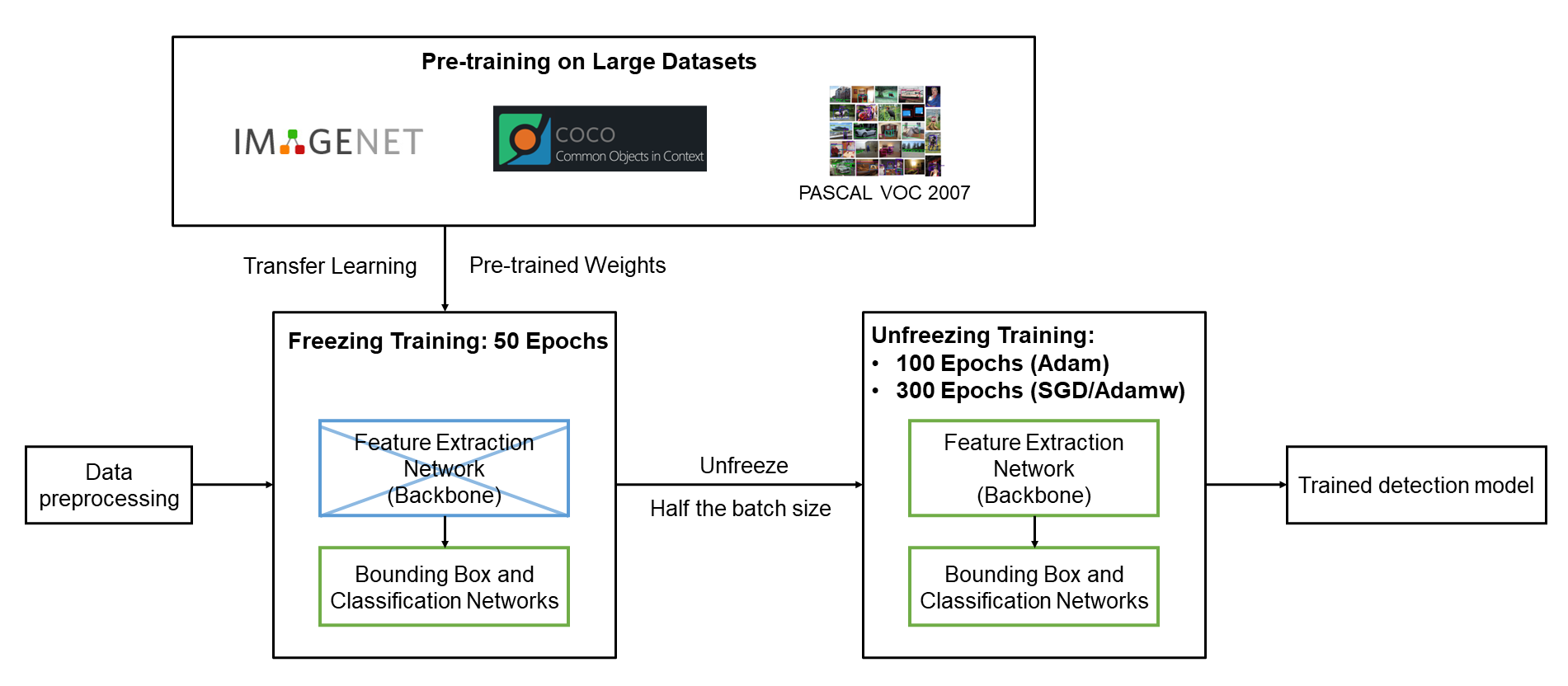}
  \caption{Two-stage supervised training process. In the freezing phase, the model's backbone is static for fine-tuning, ensuring efficient computation and swift training. This lasts for 50 epochs before transitioning to the unfreezing stage, where the entire model is adaptable, with a batch size reduced by half. For detailed training parameters, see Table \ref{tab3}. Notably, SGD and Adamw optimizer require 300 epochs while Adam optimizer requires 100 epochs in the unfreezing training. All models are fine-tuned using pre-trained weights from prominent datasets like COCO and PASCAL VOC 2007 to ensure optimal initialization.}
  \label{Supervised Training}
\end{figure*}

\begin{table*}[htbp]
\caption{high-level overview of object detection algorithms}
\renewcommand{\arraystretch}{1.2}
\begin{center}
\begin{tabular}{lccccc}
\hline
\cline{1-5}\rule{0pt}{10pt}
    \textbf{Algorithm} & \textbf{Algorithm Complexity} & \textbf{Architecture Complexity} & \textbf{Training Complexity} & \textbf{Inference Speed} &  \textbf{Robustness to Dataset Shift} \\
\midrule
    DETR  & High & High &	High &	Moderate & High	\\
    EfficientDet  & Moderate & High & Moderate & High & Moderate\\
    Faster R-CNN & High & High  & High &	Low & High\\
    RetinaNet & Moderate & Moderate & Moderate & High & Moderate\\
    YOLO  & Low & Low	 & Low &	Very high & Low\\
     SSD & Moderate & Moderate & Moderate & High & Moderate \\
     SSD-MobileNet-v2 & Low & Low & Low & Very High & Low \\
\bottomrule
\end{tabular}
\label{tab1}
\end{center}
\end{table*}

\section{Materials and methods}
\subsection{Datasets}
The dataset we utilized comprises 140 consecutive non-contrast head CT scans gathered in January 2021. These scans were acquired utilizing a 128-detector row CT scanner (SOMATOM Definition AS+, Siemens, Erlangen, Germany), with specific scanning parameters: tube voltage set to 100kV, tube current at 447mA, a field-of-view of 222mm, and a reconstruction thickness of 1mm. The demographic distribution within these 140 cases was balanced, with approximately 49\% (68 cases) being male. Age distribution within the dataset varied, ranging from 18 to 94 years.

On average, each case comprised 139 slices. Following the crucial steps of annotation and data preprocessing, a total of 619 annotated images were generated from all the slices. These images were then categorized into a training set, a validation set, and a test set following an 8:1:1 ratio. This distribution resulted in a training set consisting of 495 images, a validation set composed of 62 images, and a test set encompassing another 62 images. The distribution of four classes is 312, 286, 176, 114 instances for Right Eyes, Left Eyes, Right EAC, and Left EAC, respectively. The distinct difference in counts among these classes, especially between the Right Eyes and Left EAC, can pose challenges in model training. This dataset serves as the foundation for the proposed automated 3D reconstruction method, facilitating its validation and evaluation.

\subsection{Proposed approach overview}

Our proposed approach hinges on standardizing the CT slice data via 3D rotation prior to reconstruction. This standardization process first calls for the identification of four landmarks that determine the orbitomeatal line: bilateral eyes and bilateral external auditory canals. As depicted in Fig. \ref{f1}, we employ a deep learning-based object detection algorithm to automatically identify these four pivotal landmarks. In every slice of each sequence, the object detection algorithm is applied to spot all possible landmarks. These potential landmarks are then assessed to ascertain the most suitable ones for the respective sequence. Utilizing these landmarks, we calculate rotation angles, subsequently performing a 3D rotation for image reformatting. Ultimately, we execute the 3D reconstruction of the reformatted DICOM sequences.

To actualize automated landmark detection, we harness a supervised object detection model, which is trained on preprocessed data. Prior to training, we carry out preprocessing on the raw dataset. The preprocessed data are manually annotated by radiological technologists utilizing the annotation tool LabelImg. This labeled dataset is subsequently partitioned into a training set and a validation set, enabling the trained model to be deployed for potential landmark detection.

The exploration of the performance of object detection algorithms in the context of head CT images remains largely untapped. With the intent of identifying the most suitable algorithm for this task, we selected 12 deep learning-based object detection algorithms for a comparative analysis of both theoretical and experimental performance. As delineated in Table \ref{tab1}, we provide a high-level overview evaluating various aspects: Algorithm Complexity, Model Architecture Complexity, Training Procedure Complexity, Inference Speed, and Robustness to Dataset Shift. This systematic examination paves the way for an informed choice of the best-suited algorithm for the task at hand.

\begin{table*}[htbp]
\caption{12 object detection algorithms structures}
\renewcommand{\arraystretch}{1.2}
\begin{center}
\begin{tabular}{lccccc}
\hline
\cline{1-5}\rule{0pt}{10pt}
    \textbf{Algorithm} & \textbf{Stages} & \textbf{Backbone} & \textbf{Neck} & \textbf{Bounding box loss} & \textbf{Classification loss}\\
\midrule
   DETR\cite{bb22}  & Two & ResNet & - & L1 loss + GIoU loss	& Bipartite matching loss\\

    EfficientDet\cite{bb23}  & Two & EfficientNet & - & Smooth L1 loss & Focal loss\\

    Faster R-CNN\cite{bb24} & Two & ResNet  & - &	Smooth L1 loss & Log loss\\
    RetinaNet\cite{bb25} & One & ResNet  & - & Smooth L1 loss & Focal loss\\
    YOLOv3\cite{bb26}  & One & Darknet-53	& FPN &	Sum of squared error loss & Binary cross-entropy\\
    YOLOv4\cite{bb27} & One & CSPDarknet-53	& PANet + SSP &	CIoU loss & Binary cross-entropy\\
    YOLOv5\cite{bb28} & One & CSPDarknet-53	& PANet &	CIoU loss + Focal loss & Binary cross-entropy with Logits loss\\
    YOLOX\cite{bb29} & One & CSPDarknet-53	& PANet &	CIoU loss + Focal loss & Binary cross-entropy with Logits loss\\
    YOLOv7\cite{bb30} & One & E-ELAN	& ELAN-neck &	CIoU loss + Focal loss & Binary cross-entropy with Logits loss\\
    YOLOv8\cite{bb31} & One & CSPDarknet53 with C2f	& PANet &	CIoU loss + Distribution focal loss & VariFocal Loss\\
     SSD\cite{bb40} & One & Vgg-16 & - & Smooth L1 loss & Softmax cross entropy\\
     SSD-MobileNet-v2\cite{bb41} & One & MobileNet-v2 & - & Smooth L1 loss & Softmax cross entropy\\ 
    \bottomrule
\end{tabular}
\label{tab2}
\end{center}
\end{table*}

\subsection{Data preprocessing}
The data preprocessing involves a transition of the original DICOM data into JPEG format images, while removing the overlays. This conversion maintains the original DICOM size of 512x512 pixels. The processed images possess a resolution of 96dpi, along with an 8-bit depth and a tri-channel color scheme.

\subsection{Potential landmarks detection}

Our deep learning-based object detection algorithm identifies four pivotal landmarks: the bilateral eyes and bilateral external auditory canals. We explored 12 object detection models to aid in potential landmark detection. These models' architectures, which encompass their stages, backbone, neck, bounding box loss, and classification loss, are displayed in Table \ref{tab2}. Notably, the YOLO series models, functioning as one-stage detectors, confer the benefit of high computational speed and low computational load, albeit with a trade-off in varying accuracy degrees.

The training generally encompasses two stages: the freezing stage and the unfreezing stage. In the freezing stage, the model's backbone is immobilized, the feature extraction network remains static, and the overall network is fine-tuned. This stage offers benefits stemming from its lower computational resource demand and faster training speed. In contrast, during the unfreezing stage, the model's backbone is not frozen, leading to changes in the feature extraction network, implying that all network parameters can undergo modifications. The process initiates with 50 epochs of freezing training, subsequently advancing to unfreezing training. It is noteworthy that the batch size during the unfreezing phase is halved compared to the freezing phase.

The parameters used during training are outlined in Table \ref{tab3}. Specifically for the DETR model, we employ only the adamw optimizer to expedite training. In contrast, other models utilize both the SGD and Adam optimizers. Considering that SGD necessitates a more extended period to converge, a larger total number of epochs is set. Conversely, Adam can operate with a relatively smaller number of total epochs. Bearing in mind that training a network from scratch might yield unsatisfactory results as a result of exceedingly random weights and unremarkable feature extraction, we fine-tune each model using officially trained weights on representative large-scale datasets in the field of image recognition, such as COCO\cite{bb32}, ImageNet\cite{bb33}, and PASCAL VOC 2007\cite{pascal-voc-2007}.

\begin{figure*}[tp]
  \centering
  \includegraphics[width=\textwidth]{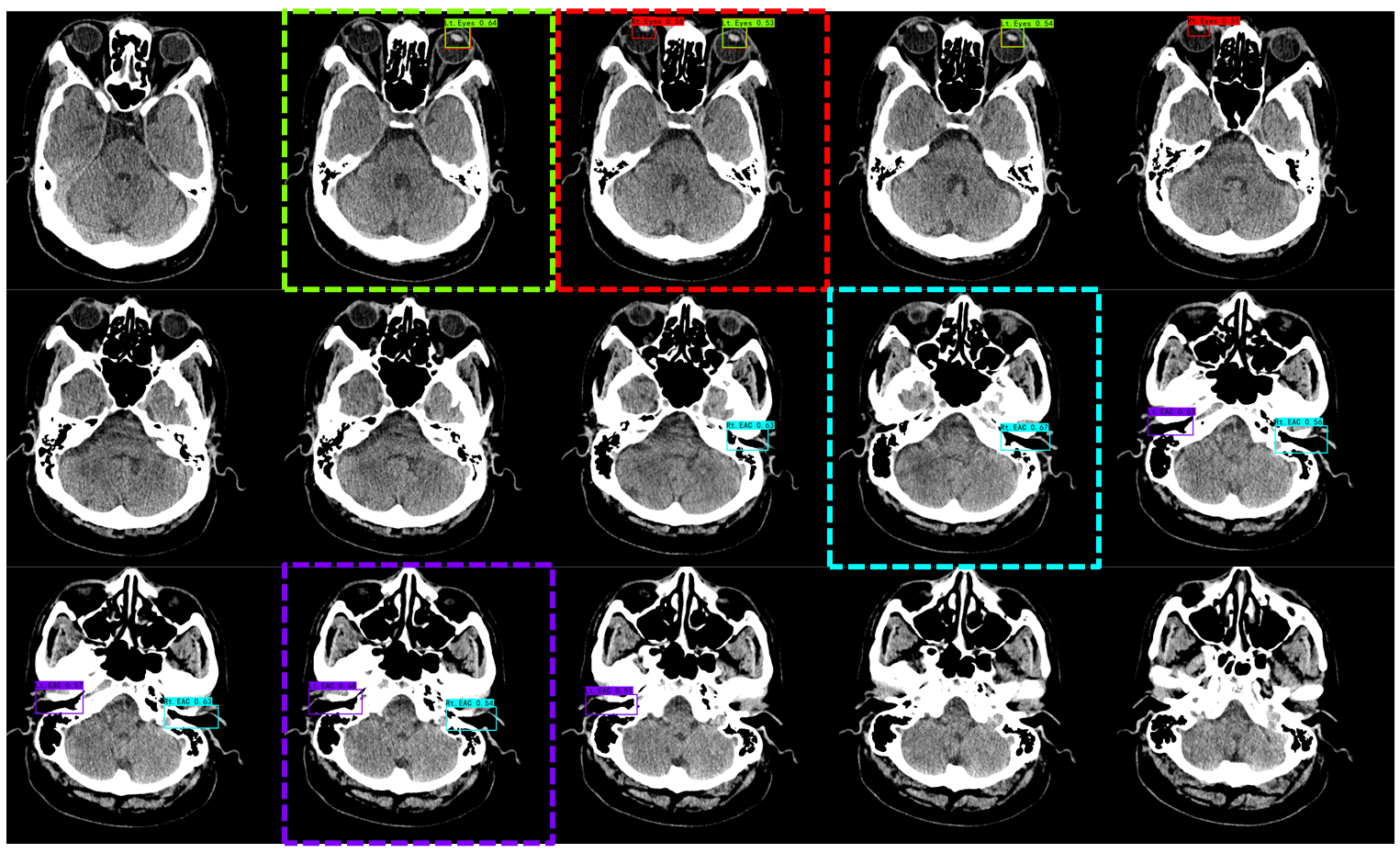}
  \caption{Landmark identification example. This is a detection result of the orbitomeatal line landmarks (bilateral eyes and bilateral EAC) on consecutive CT slices in a single case. The green, red, purple, and blue dotted boxes respectively highlight the slices where the left eye, right eye, left EAC, and right EAC were detected with the highest confidence. The indices of these selected slices are then used as the z-coordinate values for the landmarks.}
  \label{f2}
\end{figure*}

\begin{table*}[htbp]
\caption{12 object detection algorithms parameters}
\centering
\renewcommand{\arraystretch}{1.2}
\newcolumntype{C}{>{\centering\arraybackslash}X}
\newcolumntype{A}{>{\arraybackslash}X}
\begin{tabularx}{\textwidth}{ACCCCC}
\hline
\cline{1-5}\rule{0pt}{10pt}
    \textbf{Algorithm}  &\textbf{Optimizer} &\textbf{Total Epoch} & \textbf{Initial learning rate} & \textbf{Weight decay} & \textbf{Freezing batch size}\\ [0.5ex]
\hline
\vspace{0.5ex}\\ [-2ex]
   DETR  & Adamw & 300 & 0.0001 & 0.0001 & 8\\[0.5ex] 
    \hdashline
    \vspace{0.5ex}\\ [-2ex]
    \multirow{2}{*}{EfficientDet}  & Adam & 100 & 0.0003 & 0 & \multirow{2}{*}{8}\\
    & SGD & 300 & 0.01 & 0.0005\\ [0.5ex] 
    \hdashline
    \vspace{0.5ex}\\ [-2ex]
    \multirow{2}{*}{Faster R-CNN}  & Adam & 100 & 0.0001 & 0 &\multirow{2}{*}{4}\\
    & SGD & 300 & 0.01 & 0.0001\\ [0.5ex] 
    \hdashline
    \vspace{0.5ex}\\ [-2ex]
    \multirow{2}{*}{RetinaNet}  & Adam & 100 & 0.0001 & 0 & \multirow{2}{*}{8}\\
    & SGD & 300 & 0.01 & 0.0001\\[0.5ex]  
    \hdashline
    \vspace{0.5ex}\\ [-2ex]
    \multirow{2}{*}{YOLOv3}  & Adam & 100 & 0.001 & 0  & \multirow{2}{*}{16}\\
    & SGD & 300 & 0.01 & 0.0005\\[0.5ex] 
    \hdashline
    \vspace{0.5ex}\\ [-2ex]
    \multirow{2}{*}{YOLOv4 / YOLOv7 / YOLOv8}  & Adam & 100 & 0.001 & 0 & \multirow{2}{*}{8}\\
    & SGD & 300 & 0.01 & 0.0005\\[0.5ex] 
    \hdashline
    \vspace{0.5ex}\\ [-2ex]
    \multirow{2}{*}{YOLOv5 / YOLOX}  & Adam & 100 & 0.001 & 0 & \multirow{2}{*}{16}\\
    & SGD & 300 & 0.01 & 0.0005\\[0.5ex] 
    \hdashline
    \vspace{0.5ex}\\ [-2ex]
    \multirow{2}{*} {SSD / SSD-MobileNet-v2}  & Adam & 100 & 0.0006 & 0 & \multirow{2}{*}{16}\\
    & SGD & 300 & 0.002 & 0.0005\\[0.5ex] 

\bottomrule
\multicolumn{6}{p{\linewidth}}{The optimal number of epochs was primarily determined by observing early stopping criteria and monitoring the validation loss. While SGD optimizer typically demands a more extended period to reach convergence, Adam is known to work efficiently with a comparatively reduced number of epochs. For initial learning rate and weight decay, we relied on empirical benchmarks. We adopted the recommended hyperparameter values for models pre-trained on large-scale datasets such as COCO, ImageNet, and PASCAL VOC 2007. Batch size is inevitably tied to the available hardware resources. However, beyond these practical constraints, our choice was also informed by experimental outcomes, specifically focusing on the trade-off between convergence speed and accuracy.}
\end{tabularx}
\label{tab3}
\end{table*}

\subsection{Landmark identification}

In every case, the head CT image, based on its volumetric data, produces a series of potential orbitomeatal line landmark groups $G_i$, where $i\in {1,2,\cdots, N }$. Each of these groups consists of four types of landmarks: the bilateral eyes and the bilateral external auditory canals. Every landmark is equipped with a confidence score and a square bounding box, centered around the (x, y) coordinate.

To ascertain the relative $z$ position for each of the four types of landmarks, we rely on the confidence score of the bounding boxes, as expressed in the following equation:
\begin{equation}
z=\arg \max _i\left(C_i\right)
\end{equation}
Where $C_i$ stands for the bounding box's confidence score pertaining to the orbitomeatal line landmark group indexed as $i$.

Fig. \ref{f2} illustrates that, for each landmark type, the bounding box with the highest confidence score is chosen among the candidate bounding boxes. Its index serves as the relative z-coordinate. This establishes the detection box of the three-dimensional coordinates (x, y, z) as the orbitomeatal line landmark.

\begin{figure}[htbp] \centering
\includegraphics[width=0.4\textwidth,height=0.65\linewidth]{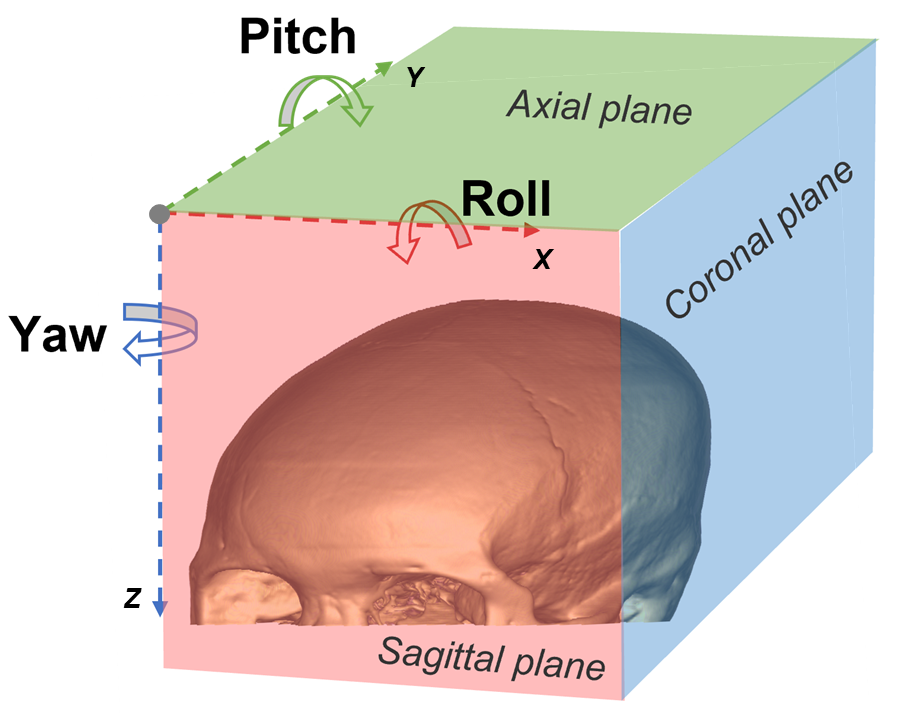}
\caption{Head CT 3D Rotation. Rotation is performed based on the Euler rotation angles (roll angle, pitch angle, and yaw angle) calculated from the orbitomeatal line landmarks coordinates.}
\label{f3}
\end{figure}

\subsection{Head CTs reformatting}
To reformat CT images, we conduct a three-dimensional rotation using Euler angles. These angles comprise a set of three distinct parameters, determining the position of a rigid body rotating around a fixed point. We compute these Euler angles based on landmarks from the bilateral eyes and external auditory canals.

Specifically, we establish a reference coordinate system, with the x-axis perpendicular to the sagittal plane, the y-axis perpendicular to the coronal plane, and the z-axis perpendicular to the axial plane.

In this system, the roll angle signifies the rotation around the x-axis. It represents the angle between the rotation vector on the sagittal plane and the y-axis. The pitch angle indicates rotation around the y-axis, i.e., the angle between the rotation vector on the coronal plane and the x-axis. The yaw angle refers to the rotation around the z-axis, or the angle between the rotation vector on the axial plane and the x-axis, as Fig. \ref{f3} illustrates.

Let the left eye landmark coordinates be $(\mathbf{x}_{Left-eye}, \mathbf{y}_{Left-eye}, \mathbf{z}_{Left-eye})$ and the right eye landmark coordinates be $(\mathbf{x}_{Right-eye}, \mathbf{y}_{Right -eye}, \mathbf{z}_{Right-eye}$). The coordinates for the left external auditory canal landmark are $(\mathbf{x}_{Left-EAC}, \mathbf{y}_{Left-EAC}, \mathbf{z}_{Left-EAC})$, and those for the right external auditory canal landmark are $(\mathbf{x}_{Right-EAC}, \mathbf{y}_{Right -EAC}, \mathbf{z}_{Right-EAC})$. Based on these landmarks, we calculate the roll angle $\boldsymbol{r}$, opting for the smaller angle value from both sides.

The roll angle is calculated as follows:

\begin{equation}
\begin{aligned}
\boldsymbol{r} = 
\min\biggl(&\arctan \left(\frac{\mathbf{z}_{Left-eye }-\mathbf{z}_{Left-EAC }}{\mathbf{y}_{Left-eye }-\mathbf{y}_{Left-EAC }}\right), \\
&\arctan \left(\frac{\mathbf{z}_{Right-eye }-\mathbf{z}_{Right-EAC }}{\mathbf{y}_{Right-eye }-\mathbf{y}_{Right-EAC }}\right)\biggr)
\end{aligned}
\end{equation}

Subsequently, we compute the pitch angle $\boldsymbol{p}$ based on the bilateral orbital and external auditory canal landmarks, again taking the smaller angle value.

The pitch angle is calculated as follows:
\begin{equation}
\begin{aligned}
\boldsymbol{p} = 
\min\biggl(&\arctan \left(\frac{\mathbf{z}_{Left-eye }-\mathbf{z}_{Right-eye }}{\mathbf{x}_{Left-eye }-\mathbf{x}_{Right-eye }}\right), \\
&\arctan \left(\frac{\mathbf{z}_{Left-EAC }-\mathbf{z}_{Right-EAC }}{\mathbf{x}_{Left-EAC  }-\mathbf{x}_{Right-EAC }}\right)\biggr)
\end{aligned}
\end{equation}

Finally, we calculate the yaw angle $\boldsymbol{y}$ based on the bilateral orbital landmarks:
\begin{equation}
\begin{aligned}
\boldsymbol{y} = 
\arctan \left(\frac{\mathbf{y}_{Left-eye }-\mathbf{y}_{Right-eye }}{\mathbf{x}_{Left-eye }-\mathbf{x}_{Right-eye }}\right)
\end{aligned}
\end{equation}

From the above calculations, we derive three rotation angles, based on which we carry out a three-dimensional rotation to obtain the reformatted DICOM format head CT images. This rotation process can be implemented with the SimpleITK library in Python, or alternatively, DICOM files can be read using the pydicom package in Python, with the rotation applied through library Scikit-image.

\subsection{Standardized reconstruction}
We undertake the three-dimensional reconstruction of the reformatted DICOM sequence utilizing the Visualization Toolkit (VTK). This task can be carried out and directly visualized with a medical image analysis platform, 3D Slicer. 

\section{Experiment and Results}\label{sec3}

\subsection{Experiment details}
The experiments were conducted in an environment comprising an Intel i7-13700k 3.4GHz CPU, an Nvidia RTX 4090 GPU with 24GB memory, and 64GB system memory. The software setup included Python 3.7, PyTorch 1.9.3, CUDA 11.6, and CUDNN 8.3.0. We employed SimpleITK 2.2.1 for 3D rotations and VTK 9.1.0 for 3D reconstructions.
\begin{table*}[htbp]
\caption{12 object detection algorithms average precision}
\centering
\renewcommand{\arraystretch}{1.2}
\newcolumntype{a}{>{\centering\arraybackslash}X}
\newcolumntype{b}{>{\arraybackslash}X}
\begin{tabularx}{\textwidth}{b a a a a a}
\hline
\cline{1-5}\rule{0pt}{10pt}
    \textbf{Algorithm} & \textbf{Left EAC AP} & \textbf{Right EAC AP} & \textbf{Left Eyes AP} & \textbf{Right Eyes AP} & \textbf{mAP}\\
\midrule
   DETR  & 0.4934 & 0.4939 & 0.4902 & 0.5686	& 0.5115\\

    EfficientDet  & \textbf{0.9881} & 0.7767 & 0.9380 & 0.9211 & \underline{0.9034}\\

    Faster R-CNN & 0.5151 & 0.6435  & 0.5704 &	0.4974 & 0.5566\\
    RetinaNet & 0.7325 & 0.5416  & 0.9724 & 0.8961 & 0.7857\\
    YOLOv3  & 0.7914 & 0.7531	& 0.9429 &	0.9125 & 0.8500\\
    YOLOv4 & 0.7091 & 0.5821	& 0.6487 &	0.5898 & 0.6324\\
    YOLOv5 & 0.5916 & 0.5750	& 0.7415 &	0.6435 & 0.6379\\
    YOLOX & 0.8458 & 0.7460	& \underline{0.9533} &	\underline{0.9376} & 0.8661\\
    YOLOv7& 0.8644 & 0.7109	& 0.9080 &	0.8857 & 0.8423\\
    YOLOv8& \underline{0.9395} & \textbf{0.8651}	& \textbf{0.9754} &	0.9366 & \textbf{0.9277}\\
    SSD& 0.8470 & \underline{0.8267}	& 0.7701 &	0.7366 & 0.7951\\
    SSD-MobileNet-v2& 0.6324 & 0.7403	& 0.9370 &	\textbf{0.9512} & 0.8152\\
    \bottomrule
    \multicolumn{6}{p{\linewidth}}{YOLOv8 exhibited the highest accuracy among all 12 object detection models, achieving a mAP of 0.9277. EfficientDet was another top-performer with an mAP of 0.9034. Each model demonstrated distinct advantages. YOLOv8 was notably robust against class imbalance, securing the highest AP within its class for right external auditory canal, left eye, and right eye.}
\end{tabularx}
\label{tab4}
\end{table*}

\begin{table*}[htbp]
\caption{12 object detection algorithms efficiency}
\centering
\renewcommand{\arraystretch}{1.2}
\newcolumntype{a}{>{\centering\arraybackslash}X}
\newcolumntype{b}{>{\arraybackslash}X}
\begin{tabularx}{\textwidth}{b a a a a a a}
\hline
\cline{1-6}\rule{0pt}{10pt}
    \textbf{Algorithm} & \textbf{mAP} & \textbf{GFLOPs} & \textbf{Total Parameters} & \textbf{MAC} & $\mathbf{PEI}$ & $\mathbf{CPEI}$\\
\midrule
   DETR & 0.5115 & 47.6260 & 36.4700 M & 28.5756 & 0.0107 & 0.0140 \\
EfficientDet & \underline{0.9034} & \underline{4.7560} & \textbf{3.8300 M} & \textbf{2.3780} & \underline{0.1905} & \textbf{0.2366} \\
Faster R-CNN & 0.5566 & 308.8130 & 136.7500 M & 200.7290 & 0.0018 & 0.0041 \\
RetinaNet & 0.7857 & 105.3140 & 36.3920 M & 63.1884 & 0.0075 & 0.0216 \\
YOLOv3 & 0.8500 & 99.3990 & 61.5400 M & 54.6695 & 0.0086 & 0.0138 \\
YOLOv4 & 0.6324 & 90.8490 & 63.9540 M & 52.6922 & 0.0070 & 0.0099 \\
YOLOv5 & 0.6379 & 73.3510 & 46.6480 M & 42.5436 & 0.0087 & 0.0137 \\
YOLOX & 0.8661 & 99.6380 & 54.1500 M & 56.7936 & 0.0087 & 0.0161 \\
YOLOv7 & 0.8423 & 67.3050 & 37.2110 M & 39.0370 & 0.0125 & 0.0226 \\
YOLOv8 & \textbf{0.9277} & 18.3380 & 11.1370 M & 9.7190 & 0.0507 & 0.0834 \\
SSD & 0.7951 & 175.6660 & 24.0130 M & 42.1600 & 0.0045 & 0.0331 \\
\makecell[l]{SSD-MobileNet-v2} & 0.8152 & \textbf{4.0060} & \underline{3.9410} M & \underline{7.9061} & \textbf{0.2035} & \underline{0.2069} \\[-8pt]
    \bottomrule
    \multicolumn{7}{p{\linewidth}}{EfficientDet outshined all models in efficiency, boasting the highest PEI and CPEI. SSD-MobileNet-v2 and YOLOv8 ranked second and third in these aspects, offering excellent choices for lightweight, high-performance detection. Other noteworthy models include YOLOX, YOLOv7, and YOLOv3, which all showed good performance with mAPs around 0.85. Given the simplicity of the YOLO architecture, these three models are also suitable for similar tasks. On the other hand, DETR and Faster R-CNN exhibited lower precision, making them unsuitable for this detection task.}
\end{tabularx}
\label{tab5}
\end{table*}

\subsection{Evaluation metrics}
We assess model performance based on factors such as accuracy, convergence, computational efficiency, and memory footprint. For accuracy, we employ metrics like mean Average Precision (mAP), F1 score, Precision, and Recall. The mAP, a prevalent metric in object detection tasks, is the mean of precision scores at varying recall levels, offering a single-figure measure of quality across these levels. We also appraise the Average Precision (AP) for the four distinct landmark classes.

As for convergence, we compute the loss on both the validation and the test sets. Computational efficiency is measured using GFLOPs (Giga Floating Point Operations Per Second), denoting the number of billion floating-point operations a system performs each second. With regard to memory usage, we assess the total parameter count of each model. This count indicates the potential memory requirement of the model, as each parameter denotes a value that the model must store and update during training.

Solely relying on the total parameter count may not present the complete picture regarding the memory efficiency of a model. Even models with similar FLOPs can have different computational speeds and memory usages. For example, group convolution consumes a large MAC and seriously affects the speed, which is not considered in FLOPs. Therefore, we incorporate the Memory Access Cost (MAC) as an efficiency evaluation metric.

To holistically observe model performance concerning accuracy, efficiency, and memory usage, we propose two indexes: the Precision Efficiency Index (PEI) and the Computational Precision Efficiency Index (CPEI):

\begin{equation}
PEI = 
\frac{mAP}{Total\ parameters}
\end{equation}
\begin{equation}
CPEI = 
\frac{mAP}{GFLOPs}
\end{equation}

In relative terms, PEI conveys the mAP per parameter present in the model, while CPEI denotes the mAP achieved per unit of computational effort.

\subsection{Detection model evaluation}

\begin{figure*}[htbp] \centering
\begin{minipage}[c]{\linewidth}
\subfloat[] {
\label{fig4a}
\includegraphics[width=0.5\linewidth,height=0.35\linewidth]{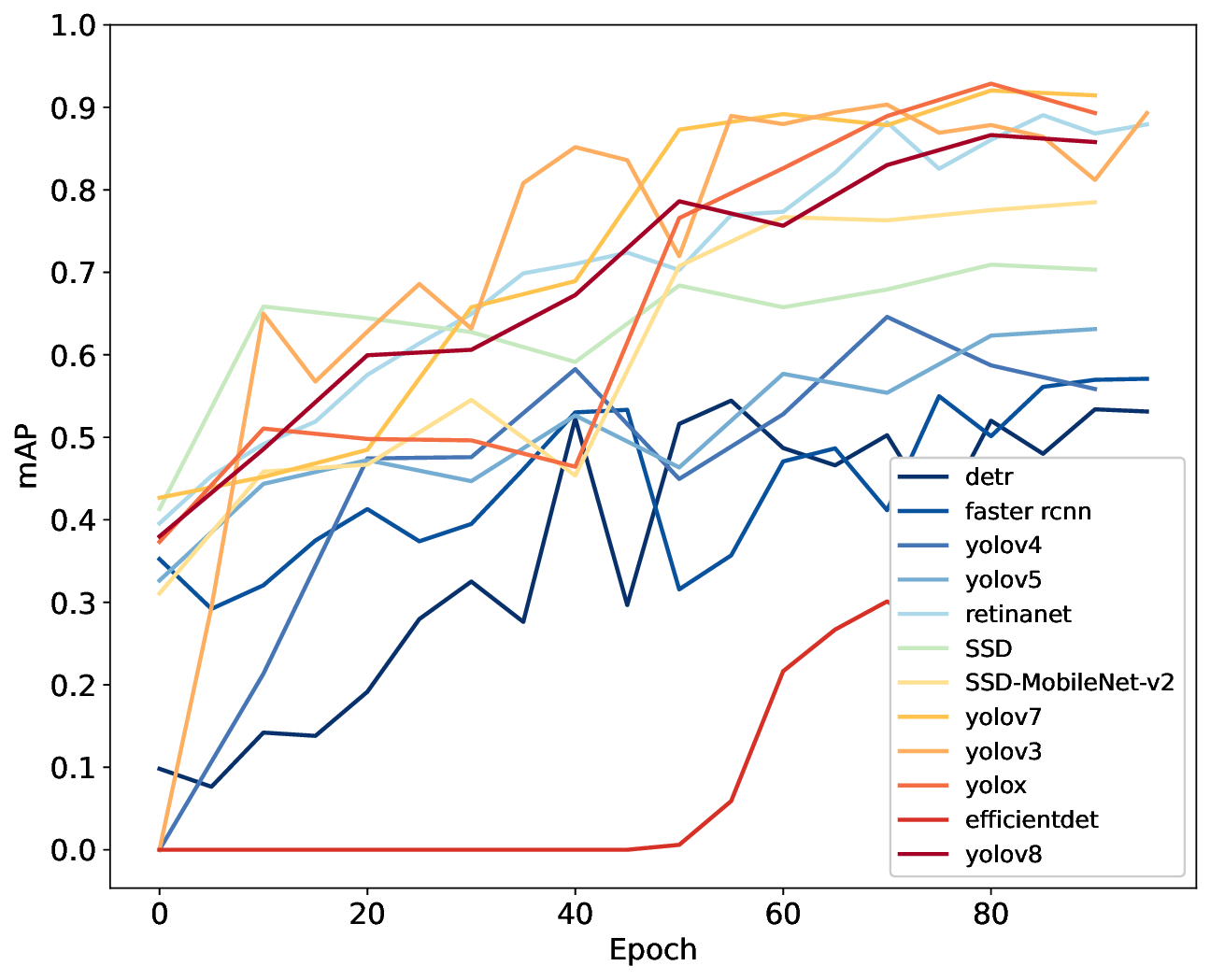}  }
\hfil
\subfloat[] {
\label{fig4b}
\includegraphics[width=0.5\linewidth,height=0.35\linewidth]{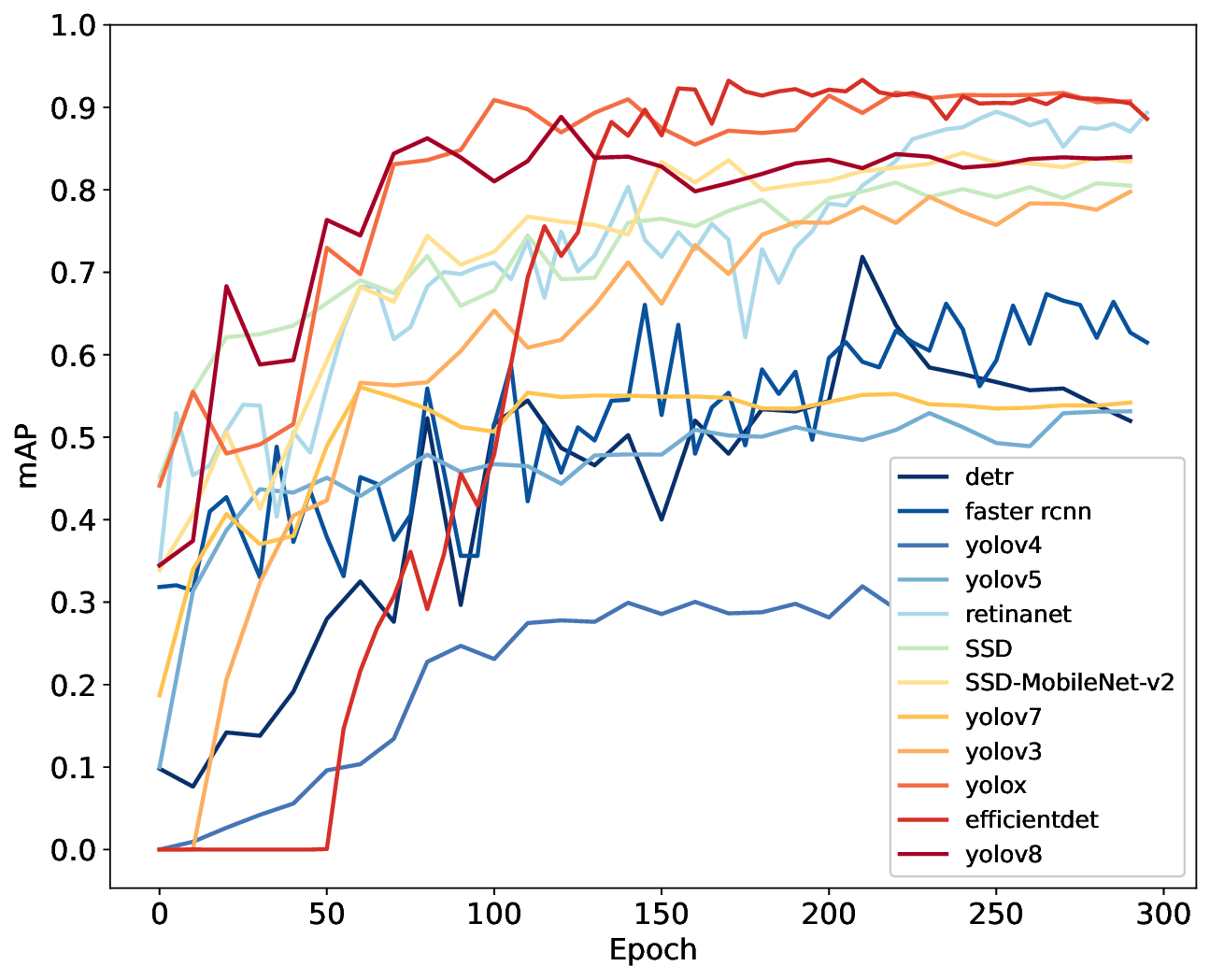}
}
\caption{Detection models mAP. (a) Adam. (b) SGD. EfficientDet's impressive accuracy relies heavily on extensive iterative epochs. EfficientDet demands 150 epochs with SGD for peak performance, compared to YOLOX's 100 epochs. EfficientDet's accuracy fell behind the other nine models, with an mAP below 0.3 after 100 epochs with Adam, yet thrives under SGD training.}
\label{fig4}
\end{minipage}
\end{figure*}

\begin{figure*}[htbp] \centering
\begin{minipage}[c]{\linewidth}
\subfloat[] {
\label{fig5a}
\includegraphics[width=0.5\linewidth,height=0.35\linewidth]{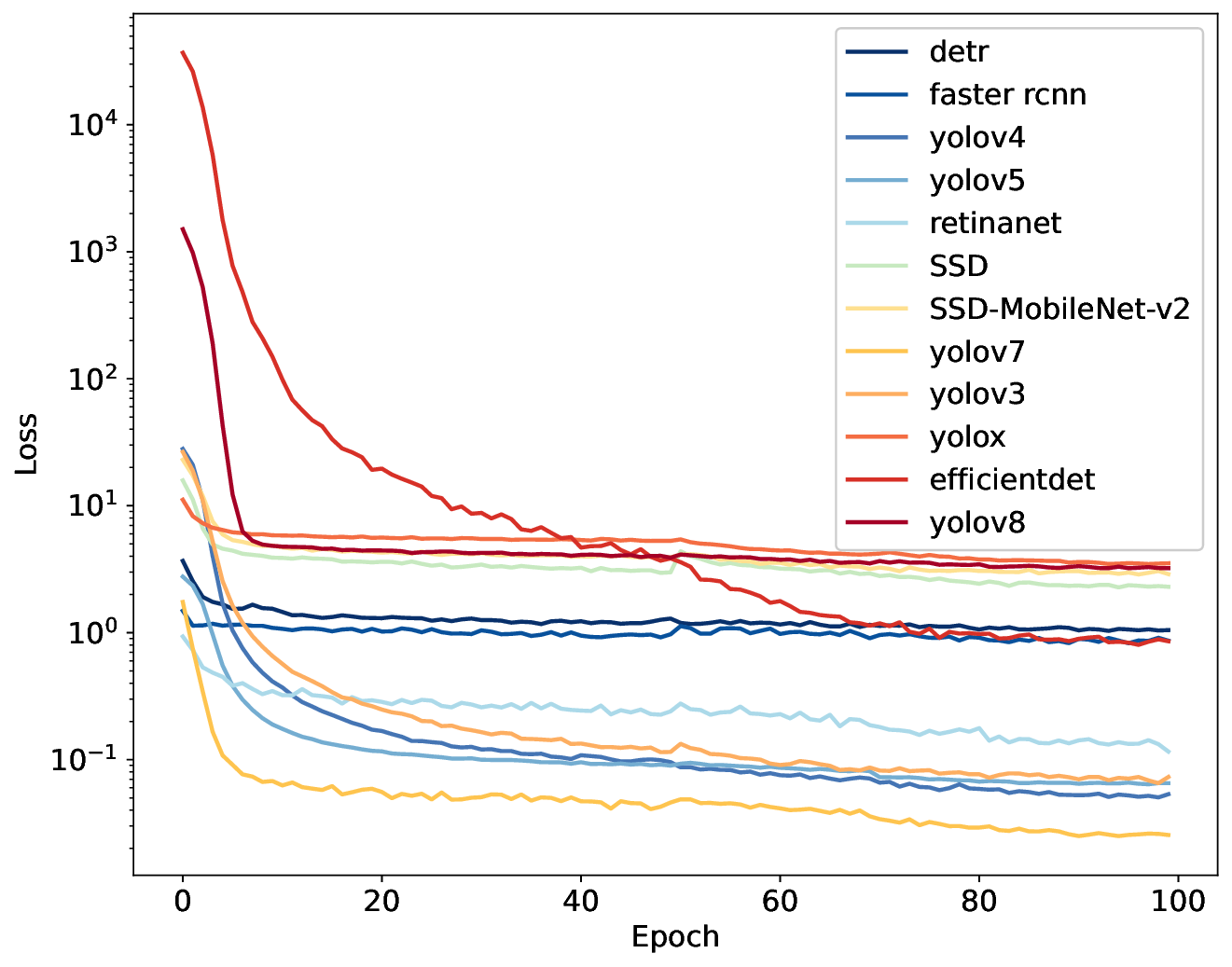}  }
\hfil
\subfloat[] {
\label{fig5b}
\includegraphics[width=0.5\linewidth,height=0.35\linewidth]{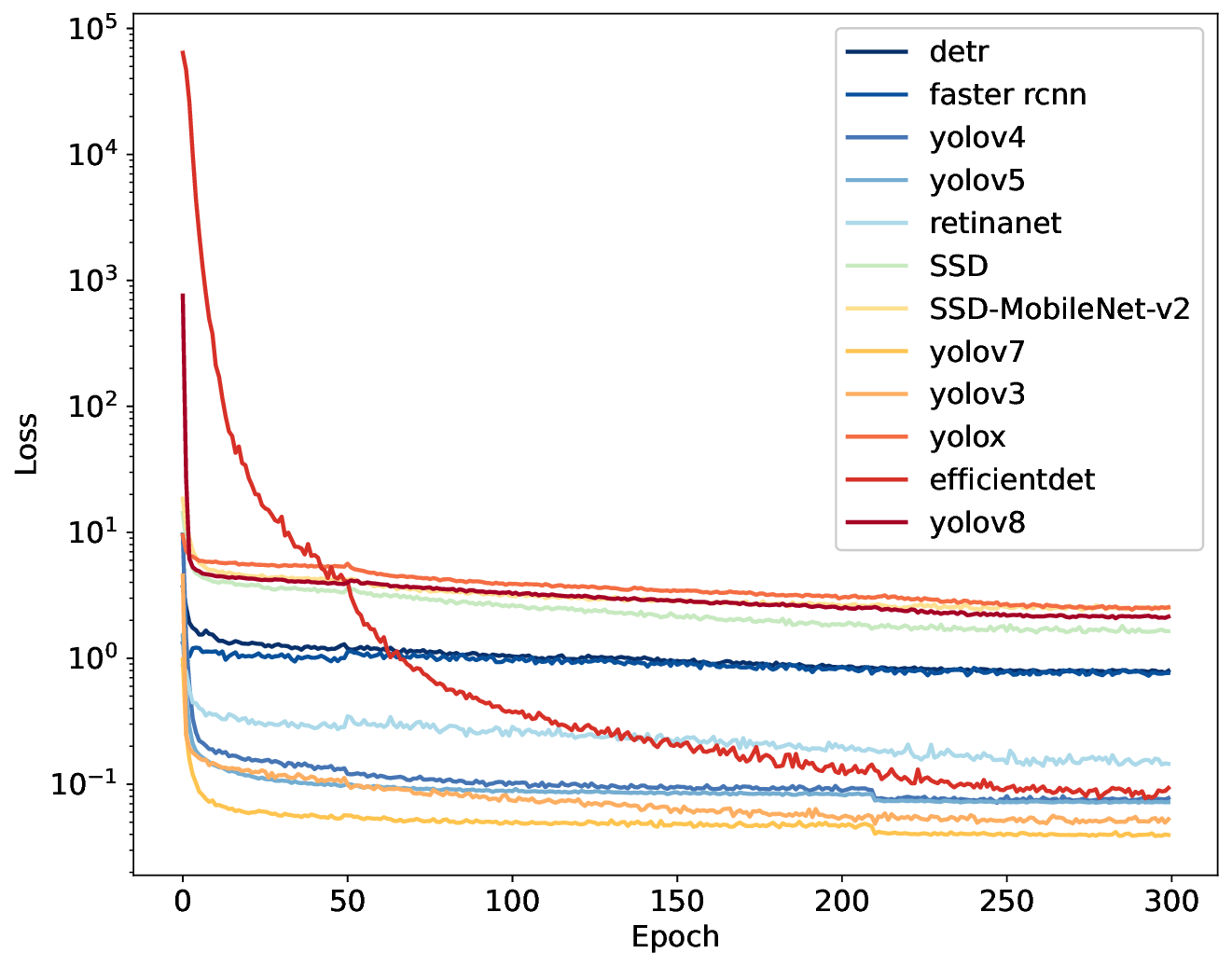}
}
\vfill
\subfloat[] {
\label{fig6a}
\includegraphics[width=0.5\linewidth,height=0.35\linewidth]{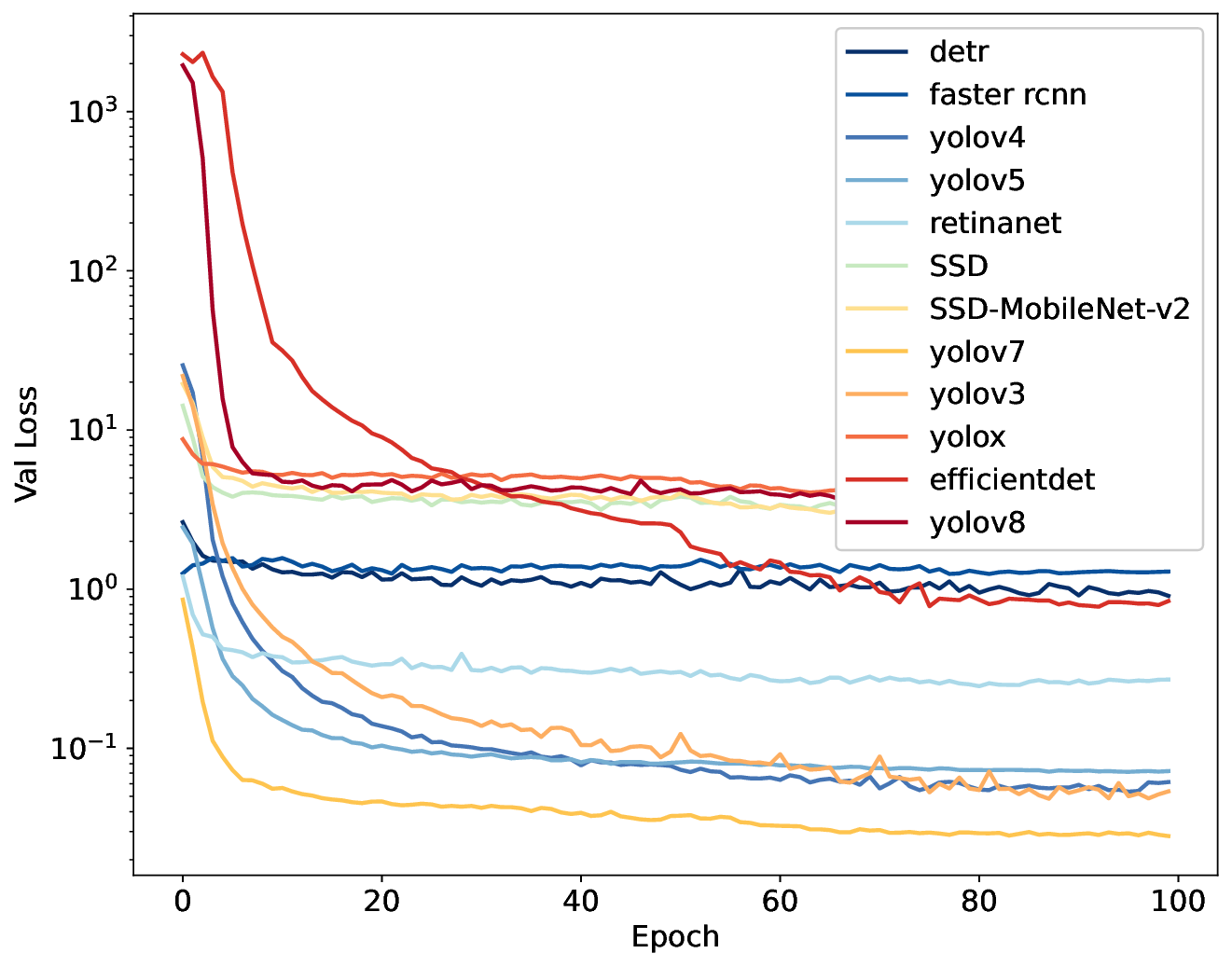}  }
\hfil
\subfloat[] {
\label{fig6b}
\includegraphics[width=0.5\linewidth,height=0.35\linewidth]{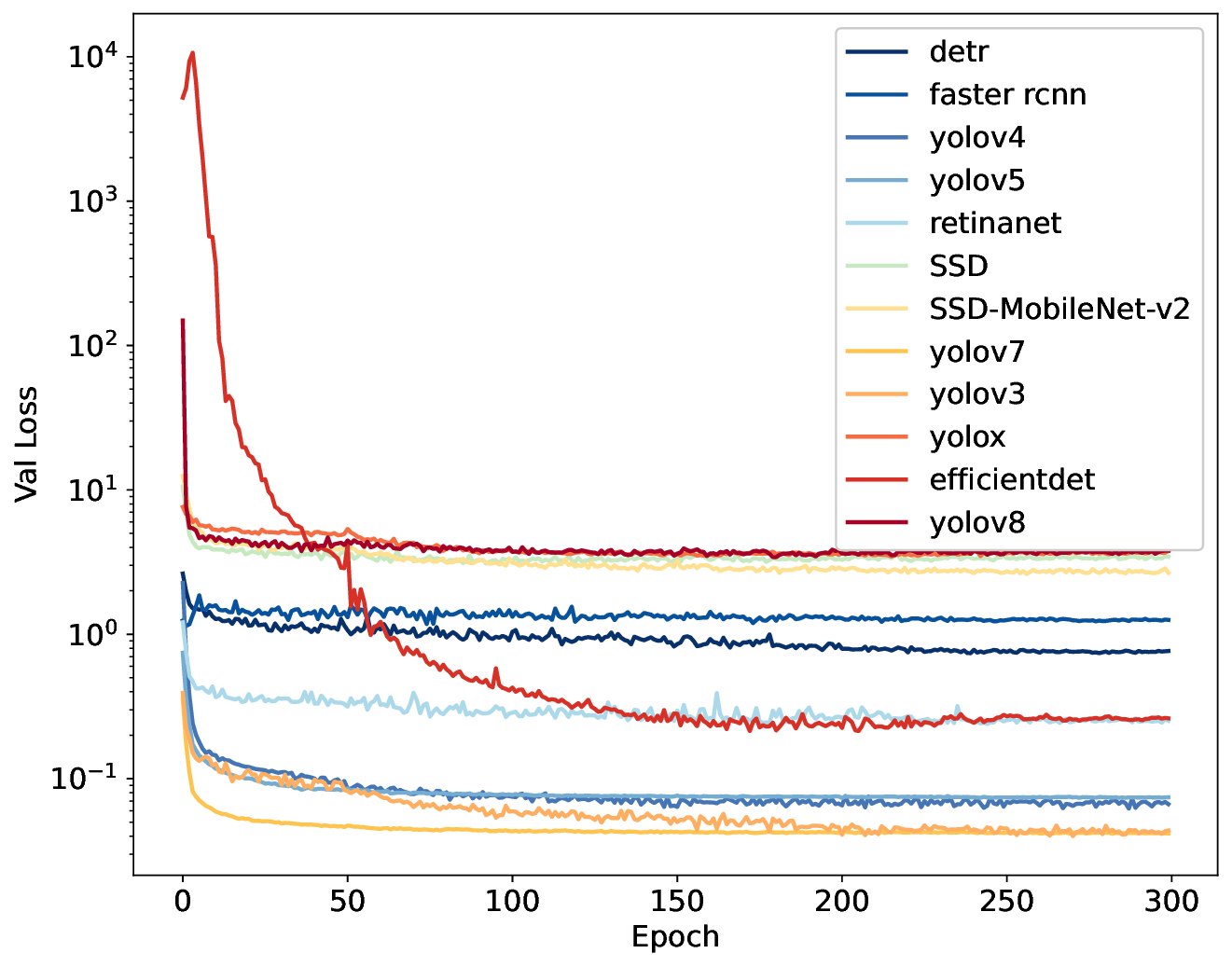}
}
\caption{Detection models loss. (a) loss with Adam. (b) loss with SGD. (c) validation loss with Adam (d) validation loss with SGD. EfficientDet is the slowest to converge among the 12 models. This underscores the potential of simple models with multiple epochs in such object detection tasks. Except for efficientdet, the other 11 models have similar convergence rates and are sufficiently learned after 10 iterations. All 12 models achieve good fit, the training loss and validation loss have converged and the difference between them is very small.}
\label{fig6}
\end{minipage}
\end{figure*}

In our experiment, the Adam optimizer outperformed SGD in terms of convergence speed, requiring 100 epochs of training compared to SGD's 300, with training data recorded every 10 epochs. 

Table \ref{tab4} reveals YOLOv8 as the most accurate among the 12 object detection models, recording a mAP of 0.9277, while EfficientDet trailed closely with 0.9034. YOLOv8 stood out for robustness against class imbalance, especially for specific classes like the right external auditory canal and both eyes. EfficientDet excelled in efficiency, as evidenced by its top PEI and CPEI scores in Table \ref{tab5}. Yet, for a balanced blend of precision and resilience, YOLOv8 is the optimal choice, accepting its minor efficiency trade-off.

EfficientDet, while precise, mandates extended training durations. Fig. \ref{fig4b} showcases it taking 150 epochs with SGD to peak, compared to YOLOX's 100 epochs. Moreover, using Adam as an optimizer for 100 epochs saw EfficientDet's mAP plummeting below 0.3 (Fig. \ref{fig4a}), yet it shined under SGD. Its slower convergence rate is evident in Fig. \ref{fig6}.

YOLOX, YOLOv7, and YOLOv3 also delivered commendable results, their mAPs hovering around 0.85, underscoring the YOLO architecture's potential. In contrast, DETR and Faster R-CNN lacked the required precision for this task.

\begin{figure*}[htbp] \centering
\begin{minipage}[c]{\linewidth}

\subfloat[] {
\label{fig11}
\includegraphics[width=\textwidth]{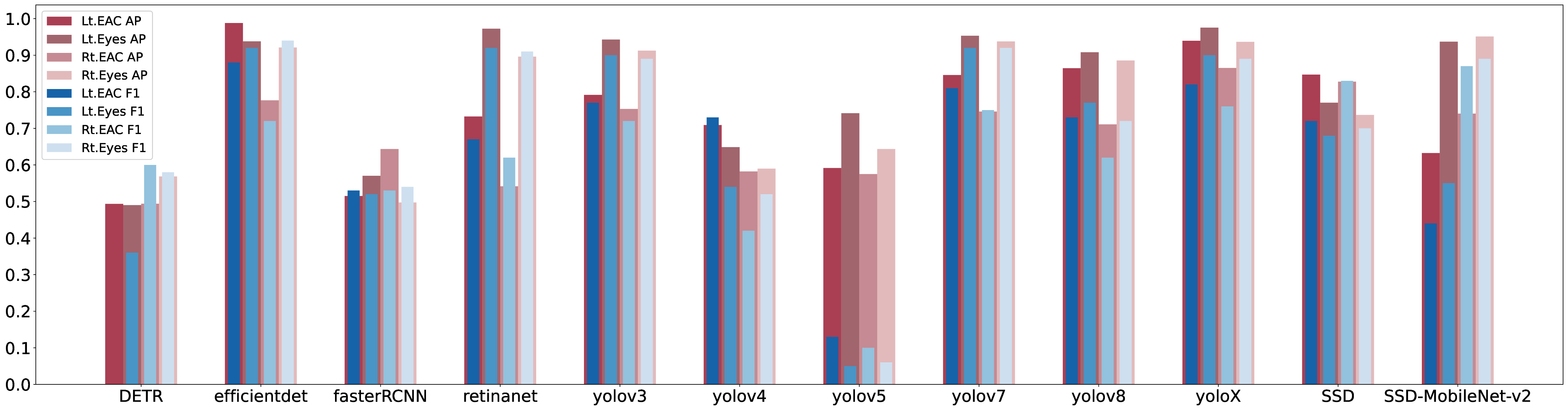}  }
\vfill

\subfloat[] {
\label{fig12a}
\includegraphics[width=0.5\linewidth,height=0.35\linewidth]{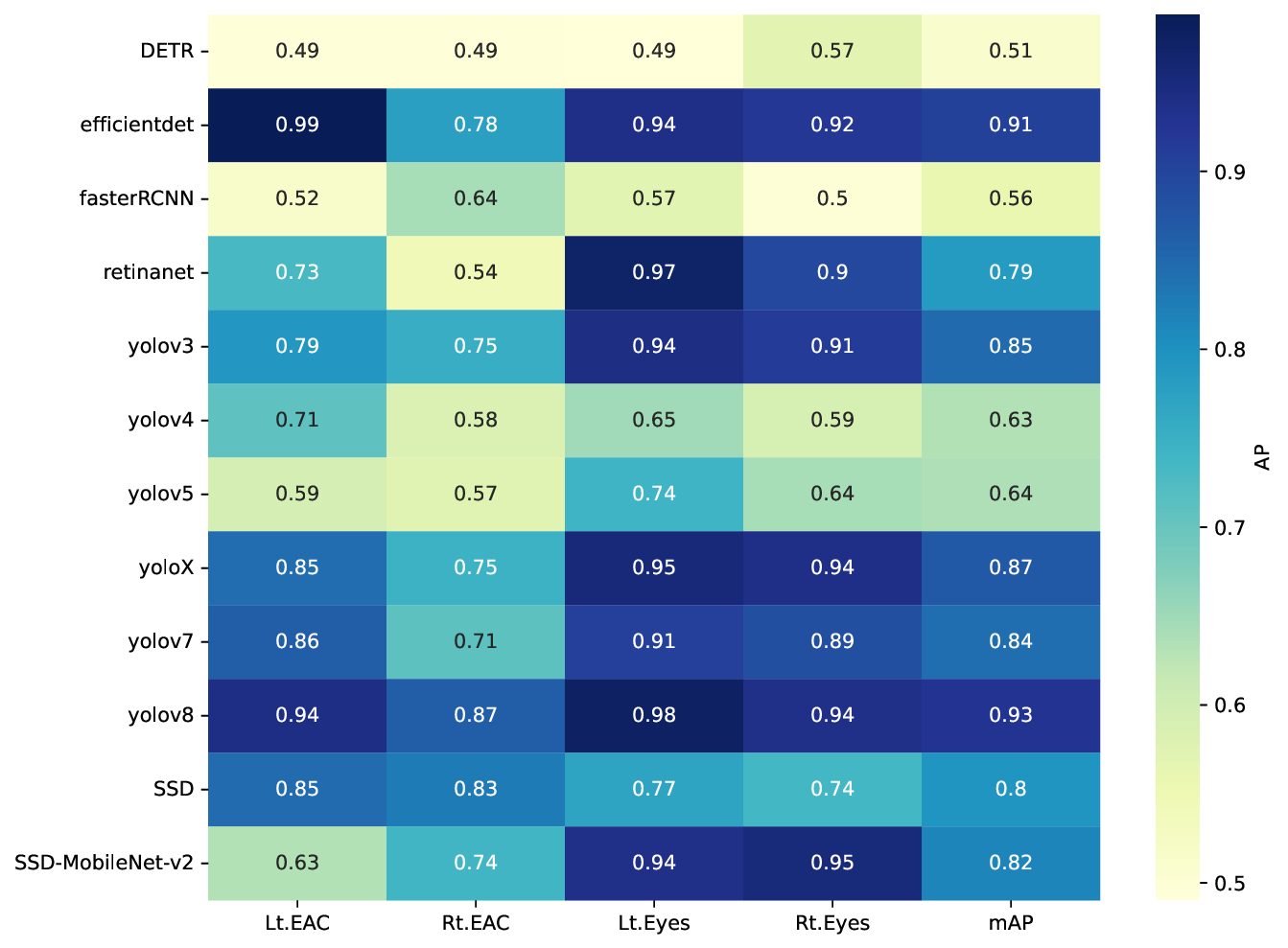}  }
\hfil
\subfloat[] {
\label{fig12b}
\includegraphics[width=0.5\linewidth,height=0.35\linewidth]{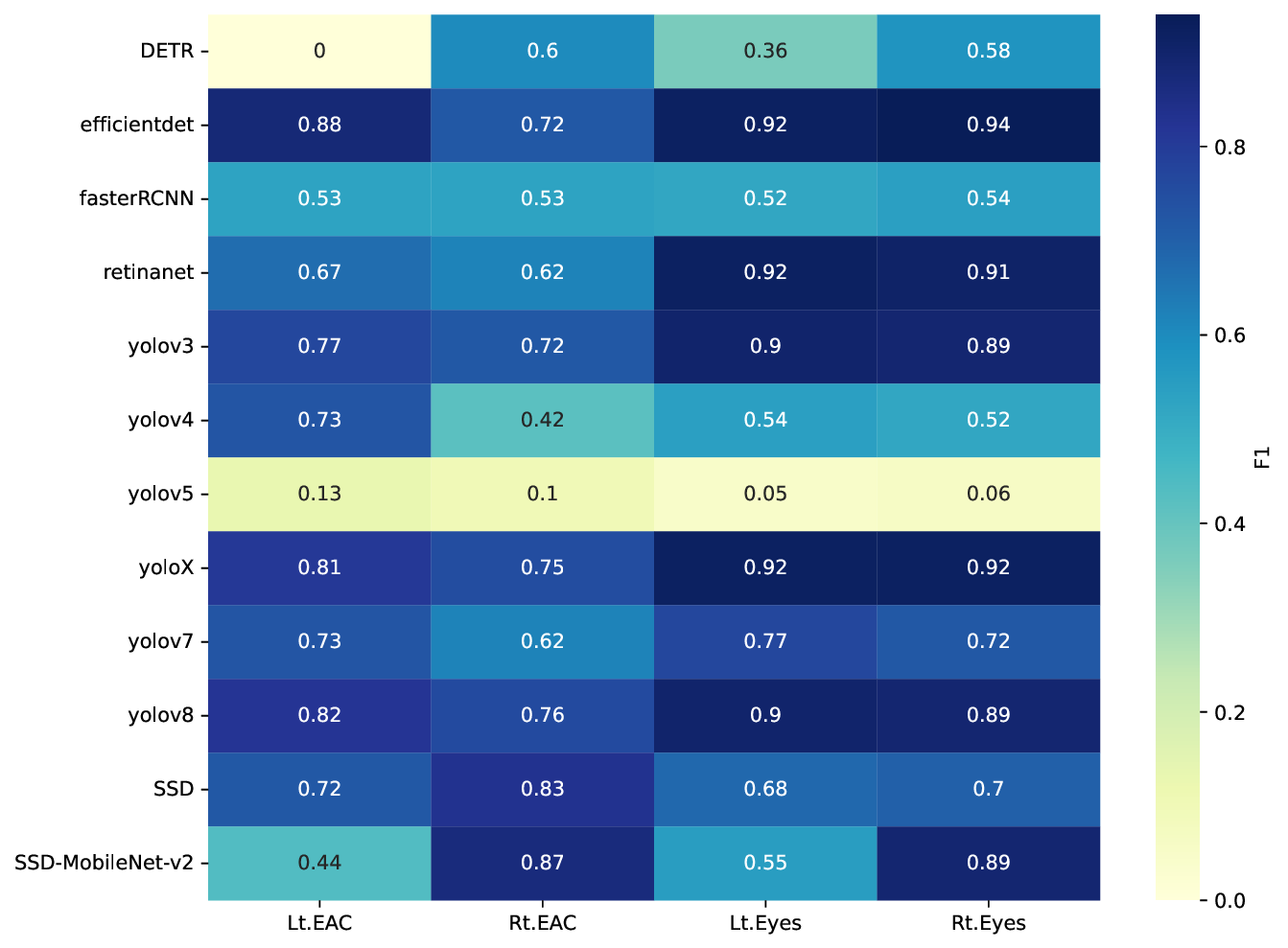}
}
\caption{(a) Results of landmarks detection for different models in terms of average precision and F1-score (score threshold = 0.5). (b) Heatmap of AP for landmarks detection. (c) Heatmap of F1-score for landmarks detection. YOLOv5 exhibits very low F1 scores across all classes, indicating a discrepancy between the model's precision and recall. This could suggest that while the model may be correctly identifying a reasonable number of objects, it could also be missing many objects or marking too many false positives, thereby leading to low F1 scores. Moreover, there are isolated instances where the F1 score exceeds the AP. This could imply that while the model's precision and recall are balanced at the specific decision threshold used for the F1 score, the model's precision may vary more across all recall levels. This suggests that the model's performance might not be consistently good for different decision thresholds.}
\label{fig12}
\end{minipage}
\end{figure*}

\begin{figure*}[htbp] \centering
\begin{minipage}[c]{\linewidth}
\subfloat[] {
\label{fig7a}
\includegraphics[width=0.5\linewidth,height=0.35\linewidth]{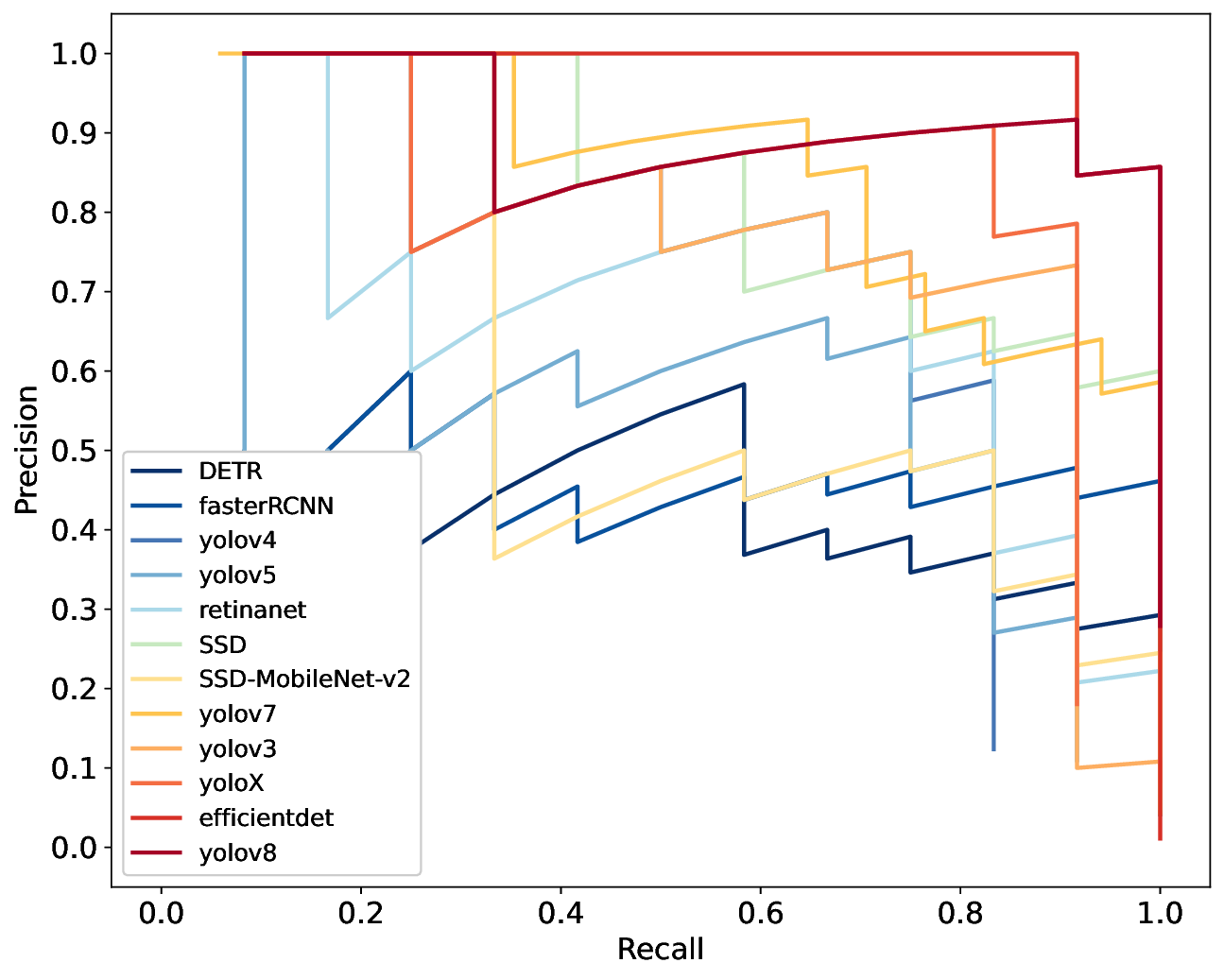}  }
\hfil
\subfloat[] {
\label{fig7b}
\includegraphics[width=0.5\linewidth,height=0.35\linewidth]{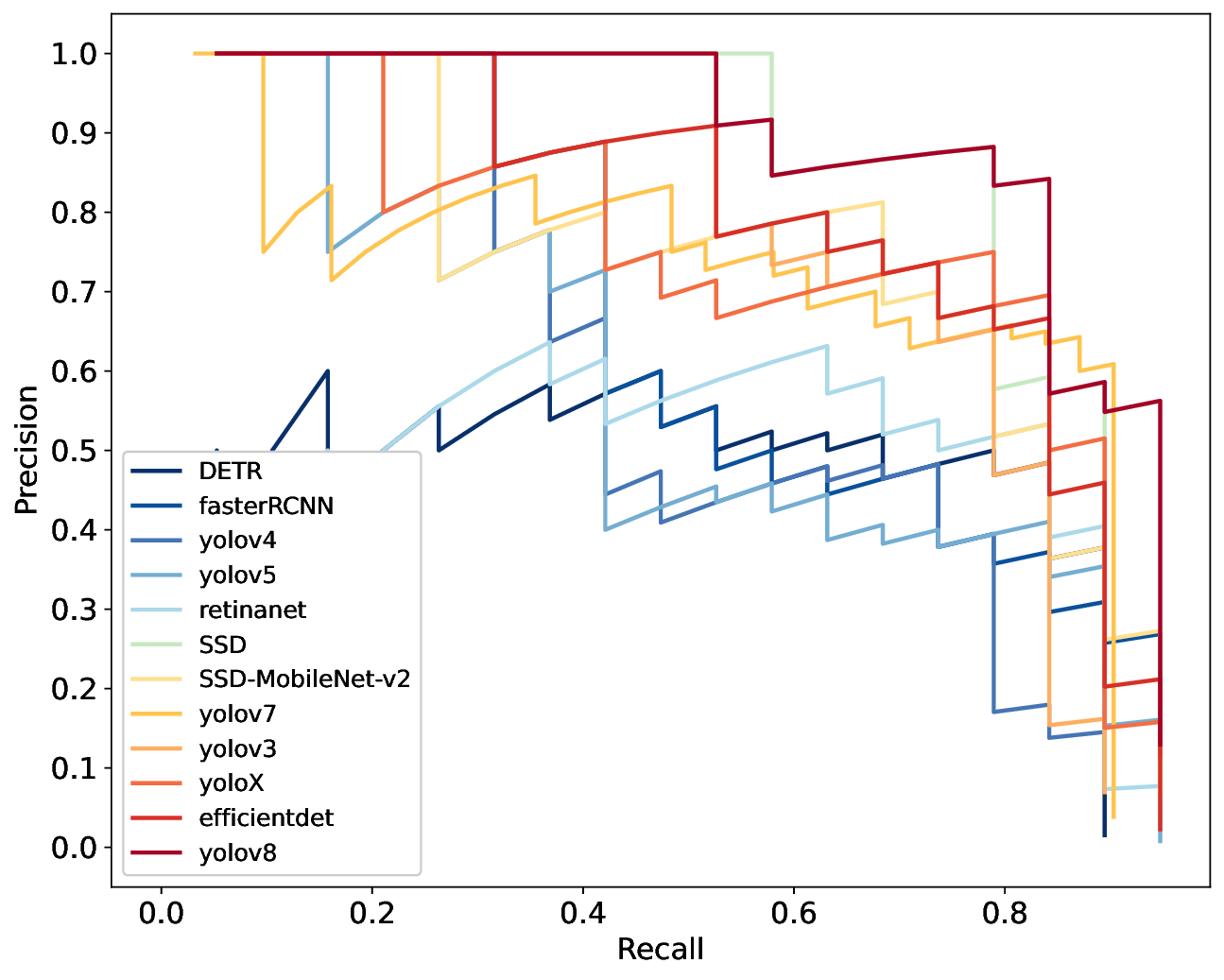}
}
\vfill
\subfloat[] {
\label{fig7c}
\includegraphics[width=0.5\linewidth,height=0.35\linewidth]{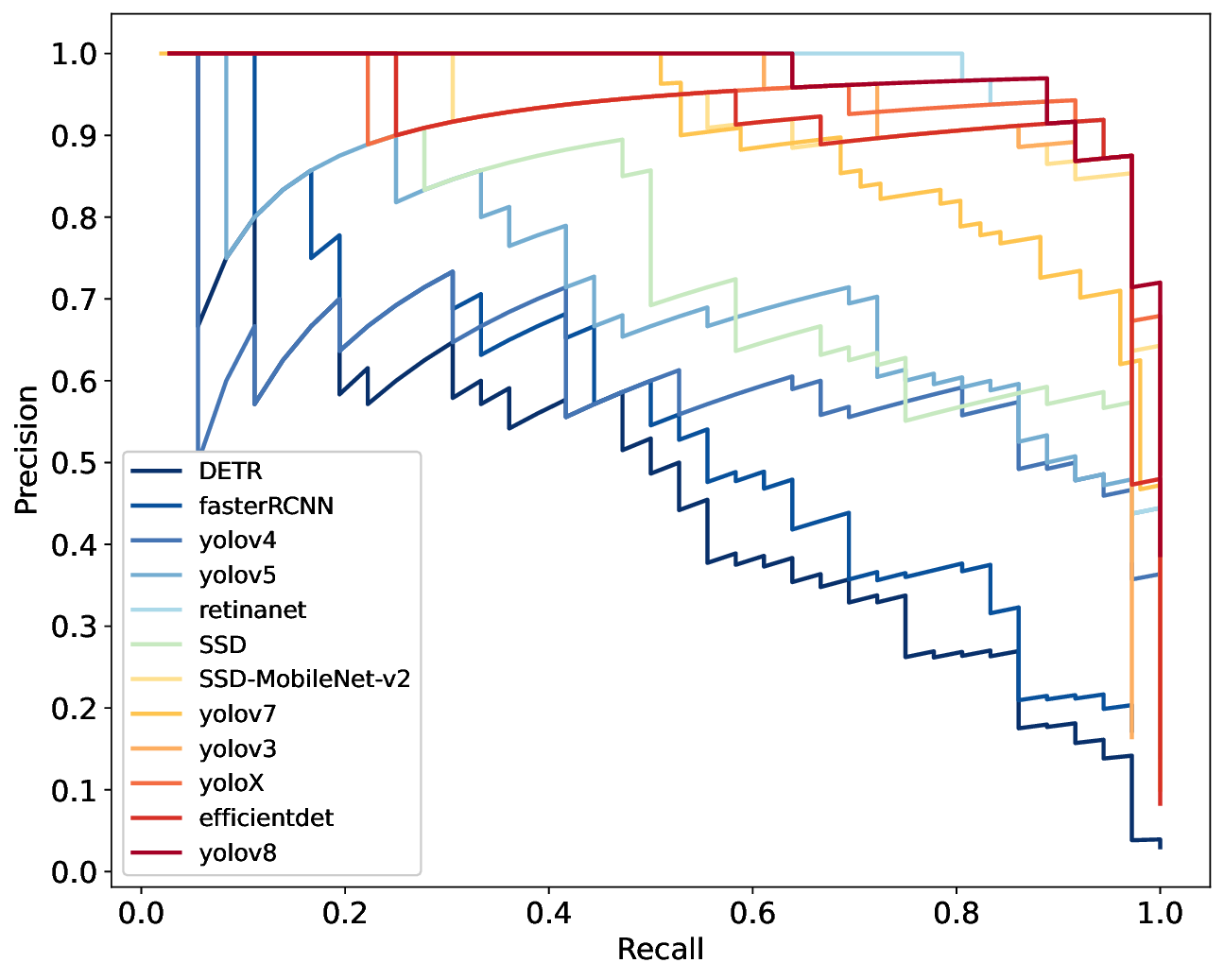}  }
\hfil
\subfloat[] {
\label{fig7d}
\includegraphics[width=0.5\linewidth,height=0.35\linewidth]{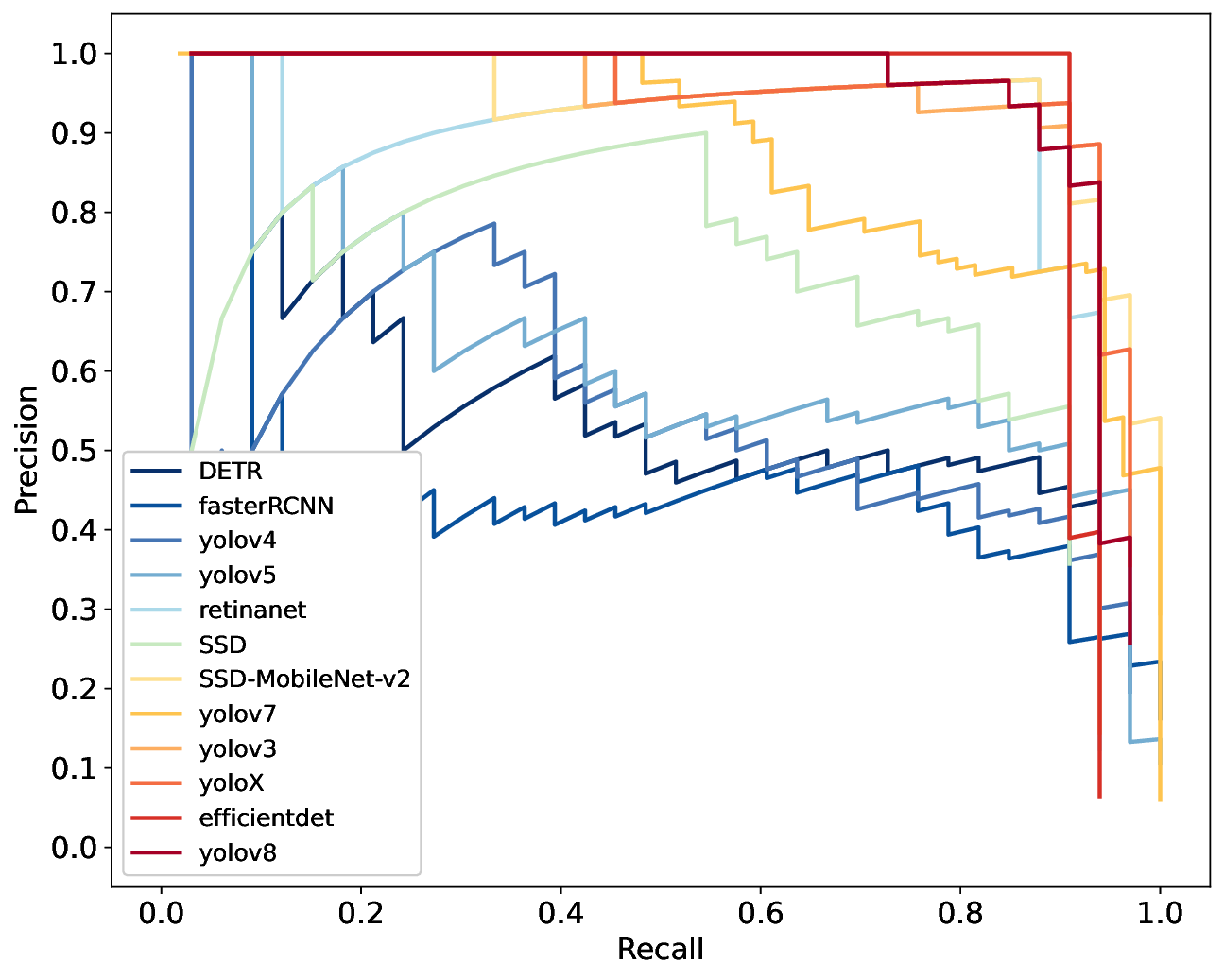}
}
\caption{Average precision for detection models. (a) Left external auditory canal. (b) Right external auditory canal. (c) Left eye. (d) Right eye. EfficientDet and YOLOv8 demonstrated superior precision across different levels of recall. Particularly when detecting samples with fewer bilateral EAC landmarks, both models managed to uphold precision while increasing recall.}
\label{fig7}
\end{minipage}
\end{figure*}

\begin{figure*}[htbp] \centering
\begin{minipage}[c]{\linewidth}
\subfloat[] {
\label{fig8a}
\includegraphics[width=0.5\linewidth,height=0.35\linewidth]{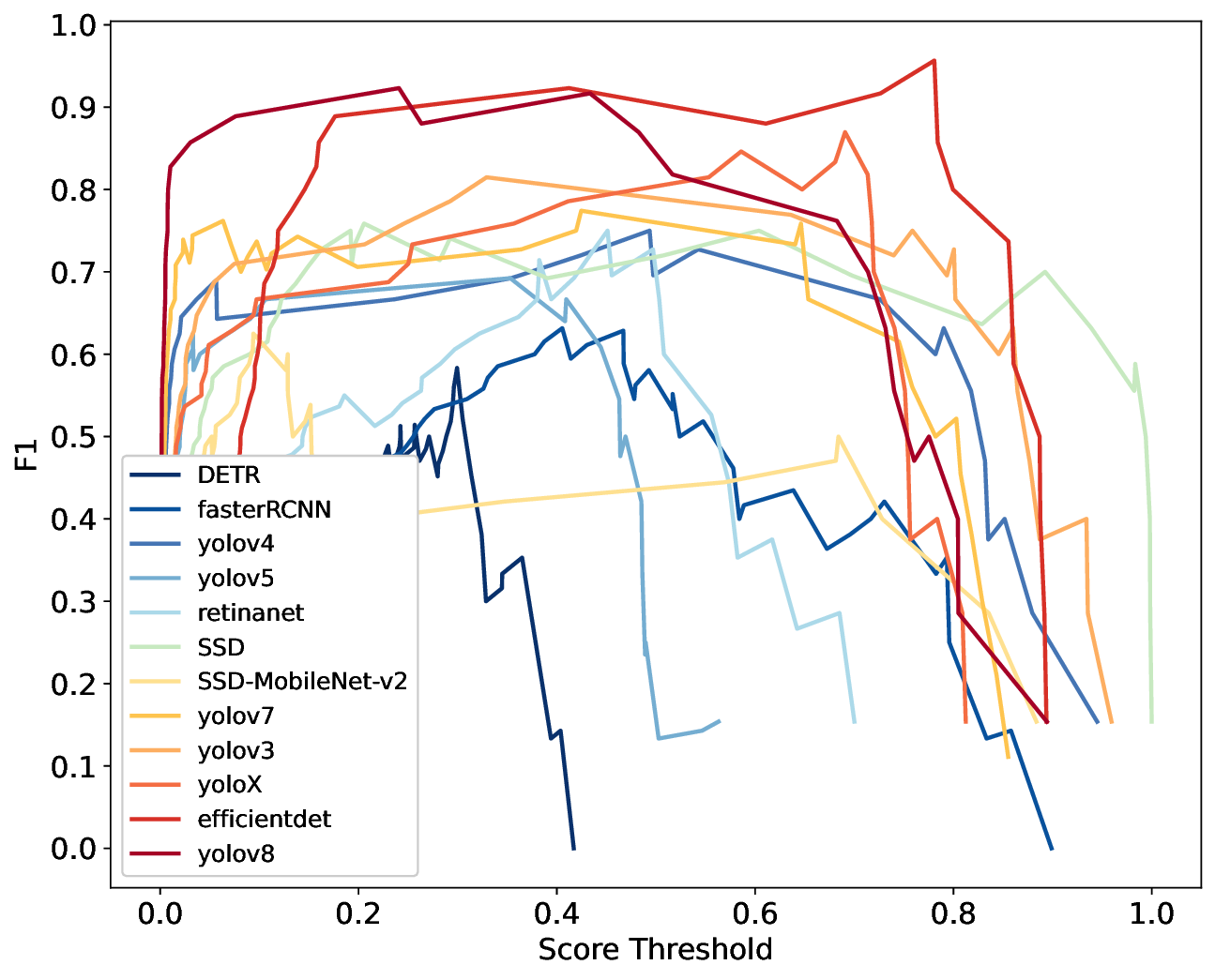}  }
\hfil
\subfloat[] {
\label{fig8b}
\includegraphics[width=0.5\linewidth,height=0.35\linewidth]{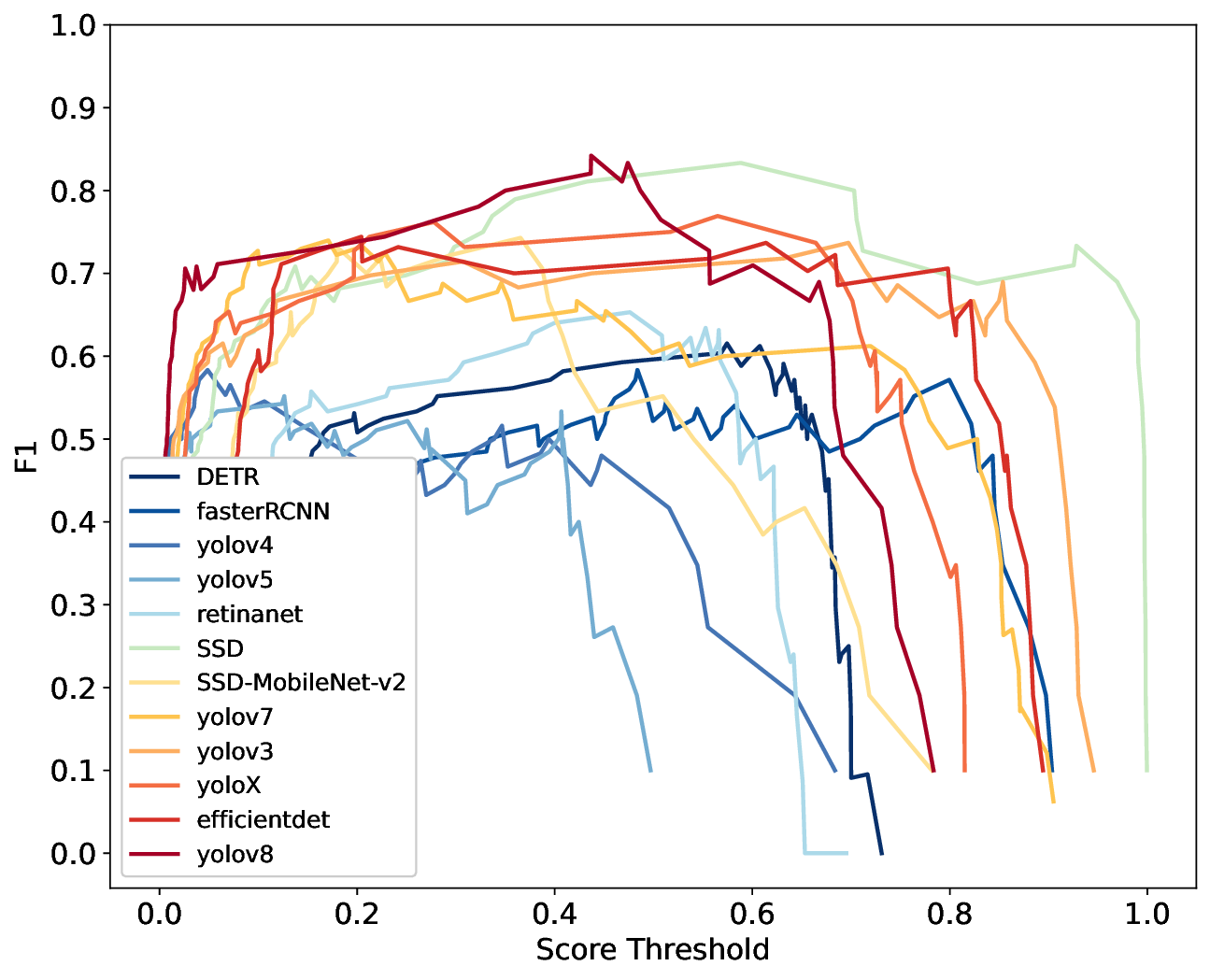}
}
\vfill
\subfloat[] {
\label{fig8c}
\includegraphics[width=0.5\linewidth,height=0.35\linewidth]{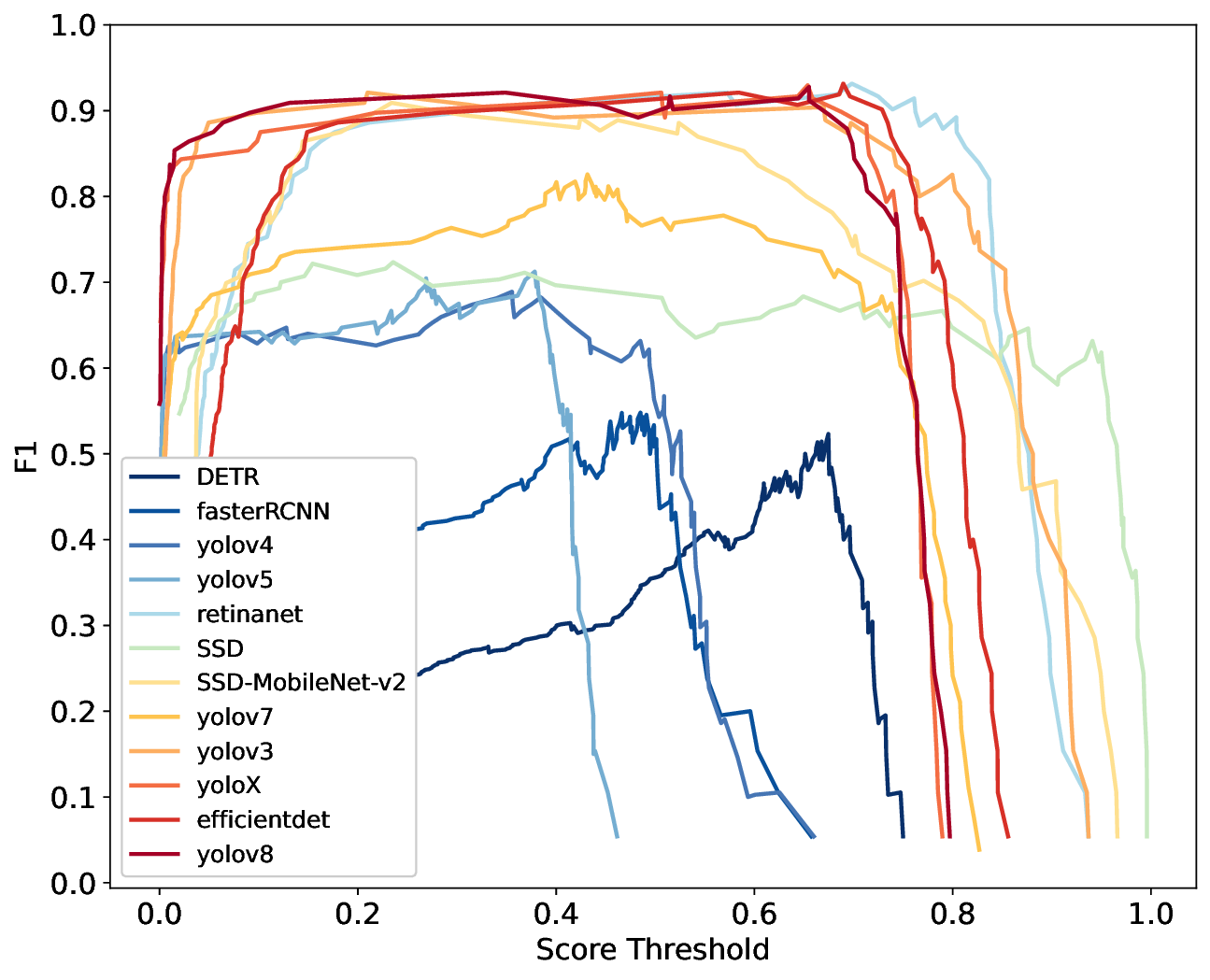}  }
\hfil
\subfloat[] {
\label{fig8d}
\includegraphics[width=0.5\linewidth,height=0.35\linewidth]{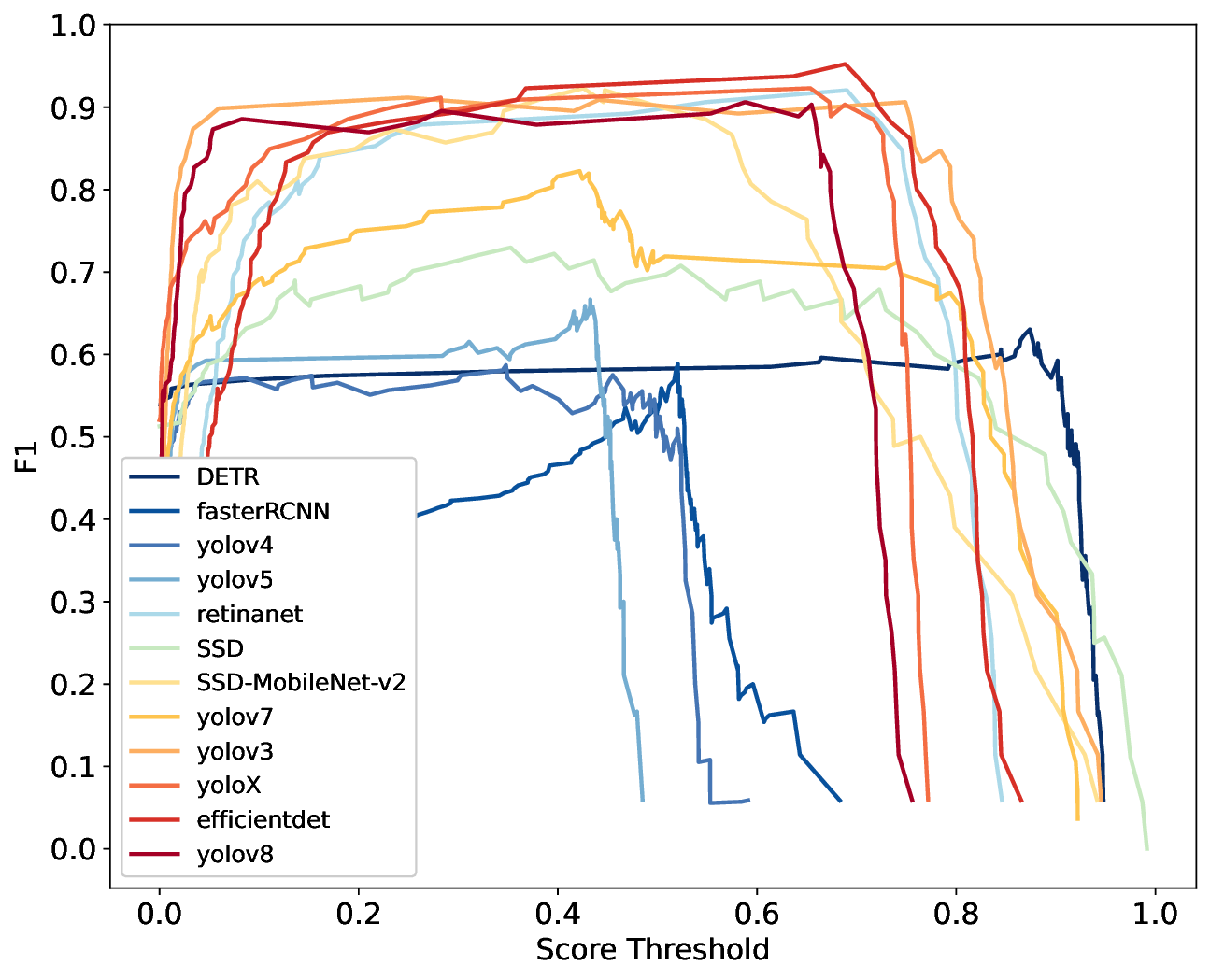}
}
\caption{F1 score for detection models. (a) Left external auditory canal. (b) Right external auditory canal. (c) Left eye. (d) Right eye. In terms of F1 scores, YOLOv8 consistently performed well across all four landmark detections. YOLOv8 achieved peak F1 scores at a lower threshold, denoting high F1 scores without a stringent threshold. None of the models exhibited consistent performance across thresholds for bilateral EACs, likely due to smaller sample sizes. Conversely, for bilateral eyes where the sample size was larger, YOLOv3, YOLOX, and YOLOv8 maintained high F1 scores across a wide threshold range, reflecting more robust models that perform well irrespective of the precise threshold.}
\label{fig8}
\end{minipage}
\end{figure*}

\begin{figure*}[htbp] \centering
\begin{minipage}[c]{\linewidth}
\subfloat[] {
\label{fig9a}
\includegraphics[width=0.5\linewidth,height=0.35\linewidth]{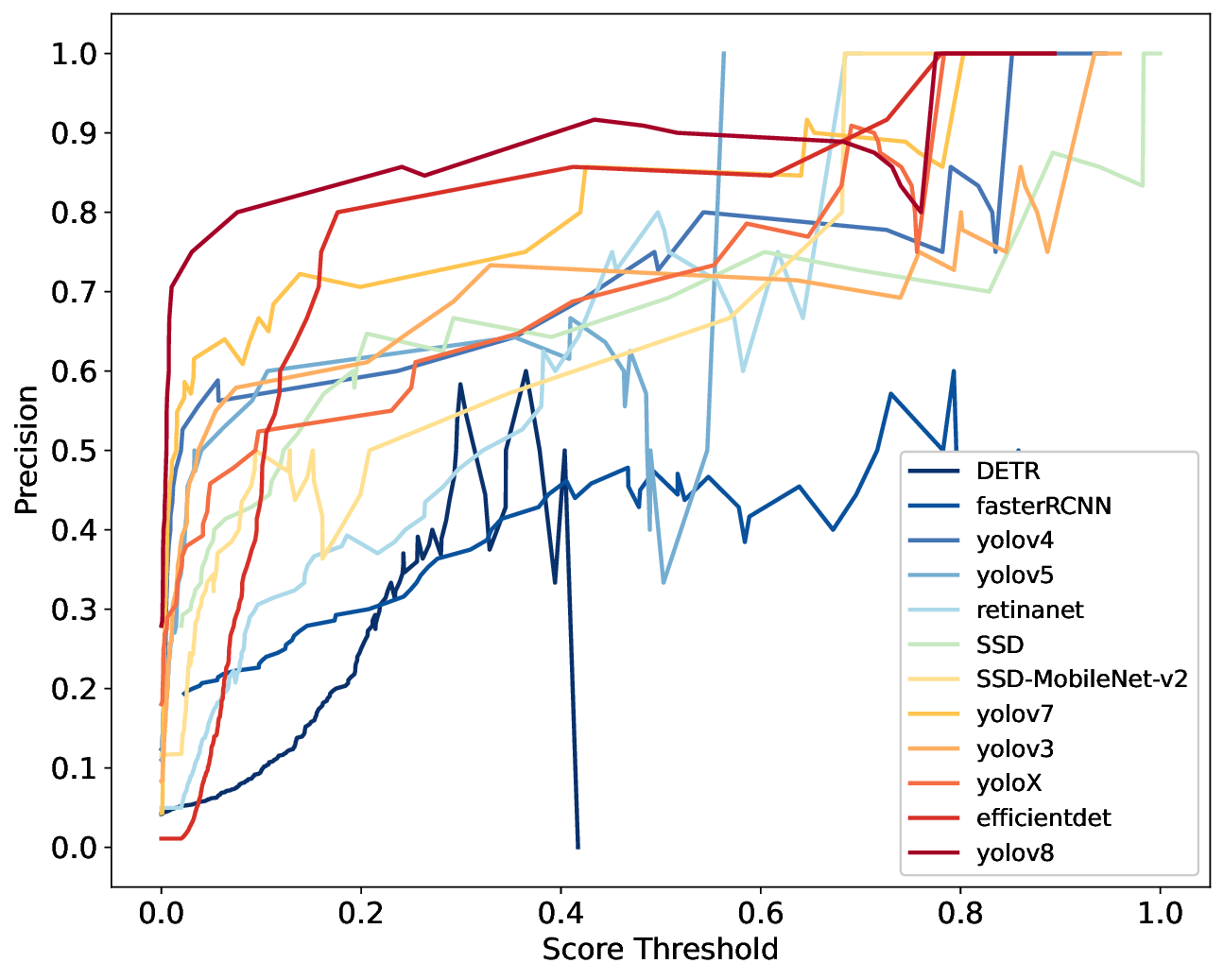}  }
\hfil
\subfloat[] {
\label{fig9b}
\includegraphics[width=0.5\linewidth,height=0.35\linewidth]{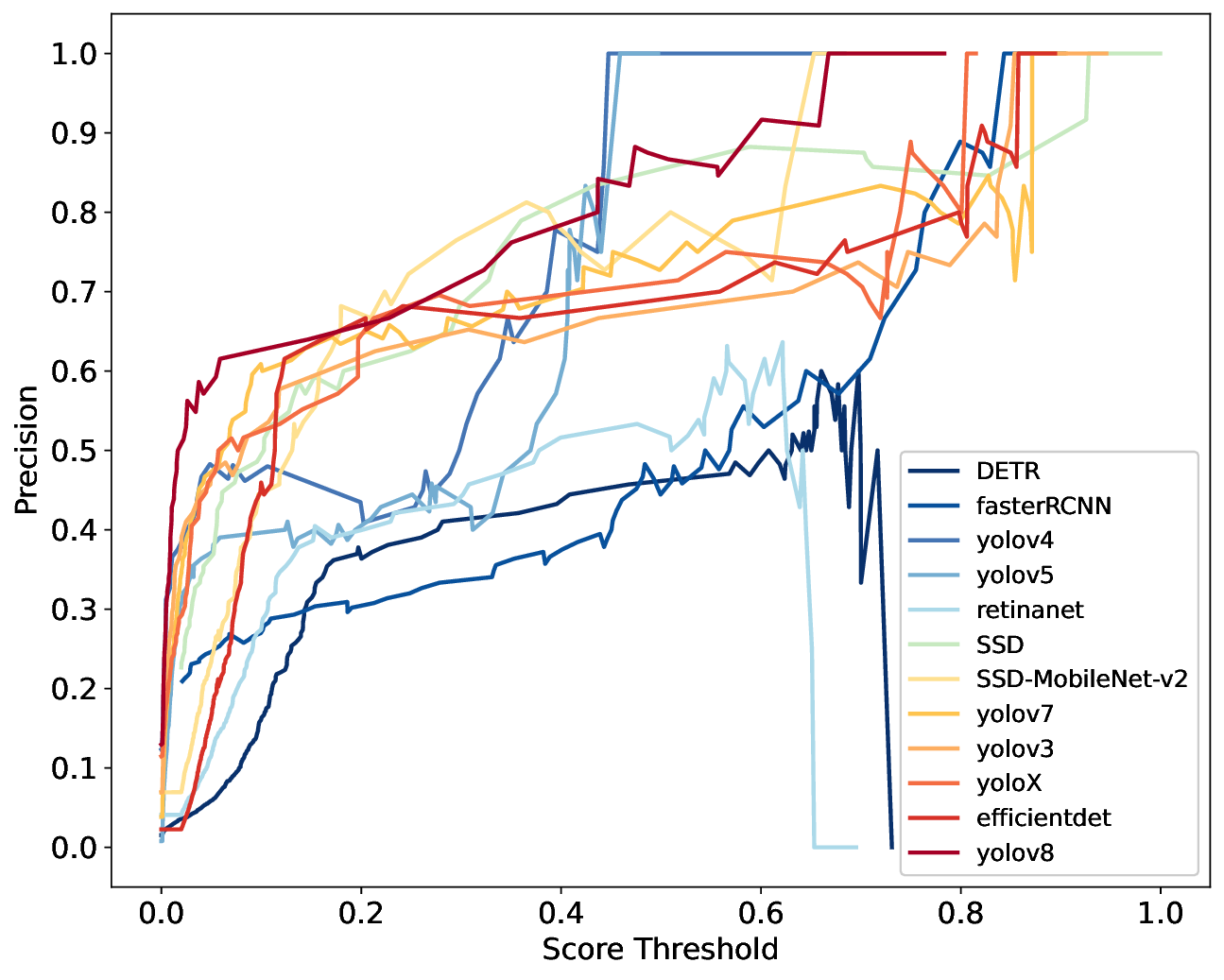}
}
\vfill
\subfloat[] {
\label{fig9c}
\includegraphics[width=0.5\linewidth,height=0.35\linewidth]{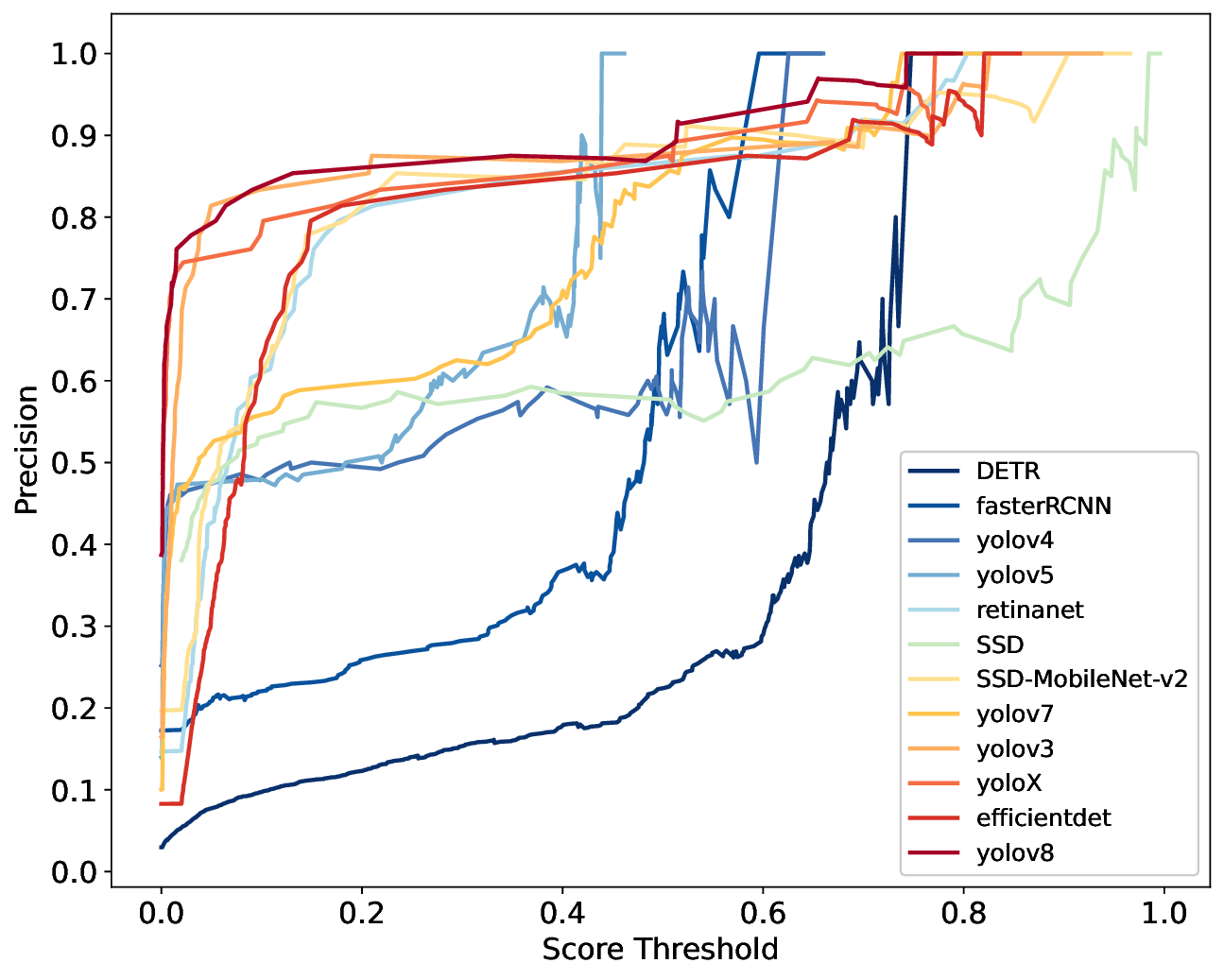}  }
\hfil
\subfloat[] {
\label{fig9d}
\includegraphics[width=0.5\linewidth,height=0.35\linewidth]{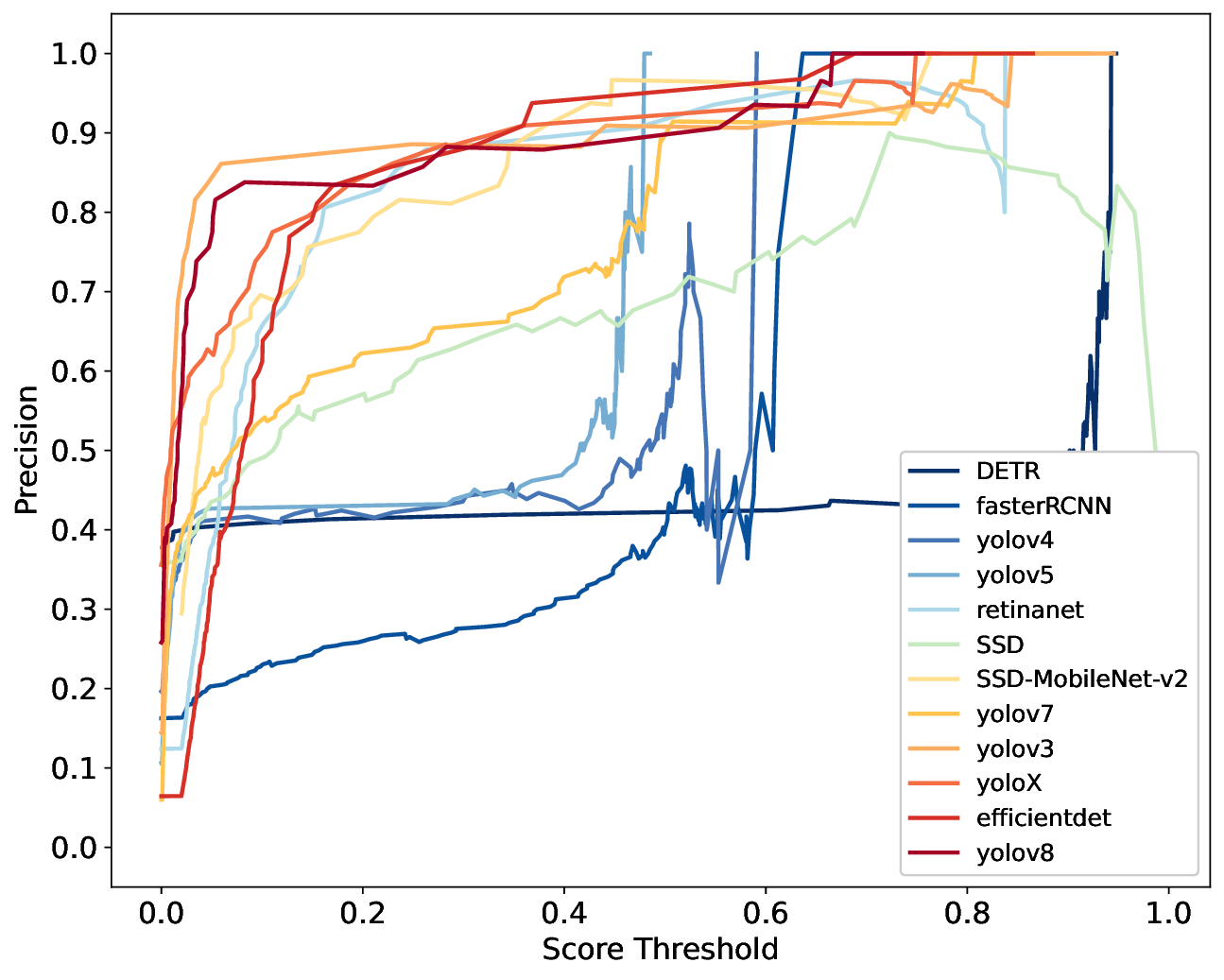}
}
\caption{Precision for detection models. (a) Left external auditory canal. (b) Right external auditory canal. (c) Left eye. (d) Right eye. In the predictions across the four classes, DETR, Faster RCNN, YOLOv4, and YOLOv5 exhibited several sharp drops or spikes in precision, rendering them less reliable. YOLOv8, on the other hand, consistently outperformed its counterparts across various score thresholds maintaining high precisions at high confidence levels, thus emerging as the most robust model. EfficientDet closely follows in terms of performance.}
\label{fig9}
\end{minipage}
\end{figure*}

Fig. \ref{fig7} highlights the impeccable precision of EfficientDet and YOLOv8 at various recall levels, more so with limited bilateral EAC landmarks. YOLOv8 consistently scored high in F1 scores for all four landmark detections, evident in Fig. \ref{fig8}. For bilateral eyes with ample samples, YOLOv3, YOLOX, and YOLOv8 showcased broad threshold range robustness, as reinforced by Fig. \ref{fig9c} and Fig. \ref{fig9d}. Fig. \ref{fig11} captures the performance nuances when the score threshold is at 0.5. Notably, YOLOv5 had significantly low F1 scores, hinting at possible model inefficiencies. Cases of F1 scores surpassing AP indicate potential performance inconsistencies across decision thresholds.


\begin{figure*}[htbp] \centering
\begin{minipage}[c]{\linewidth}
\subfloat[] {
\label{fig10a}
\includegraphics[width=0.5\linewidth,height=0.35\linewidth]{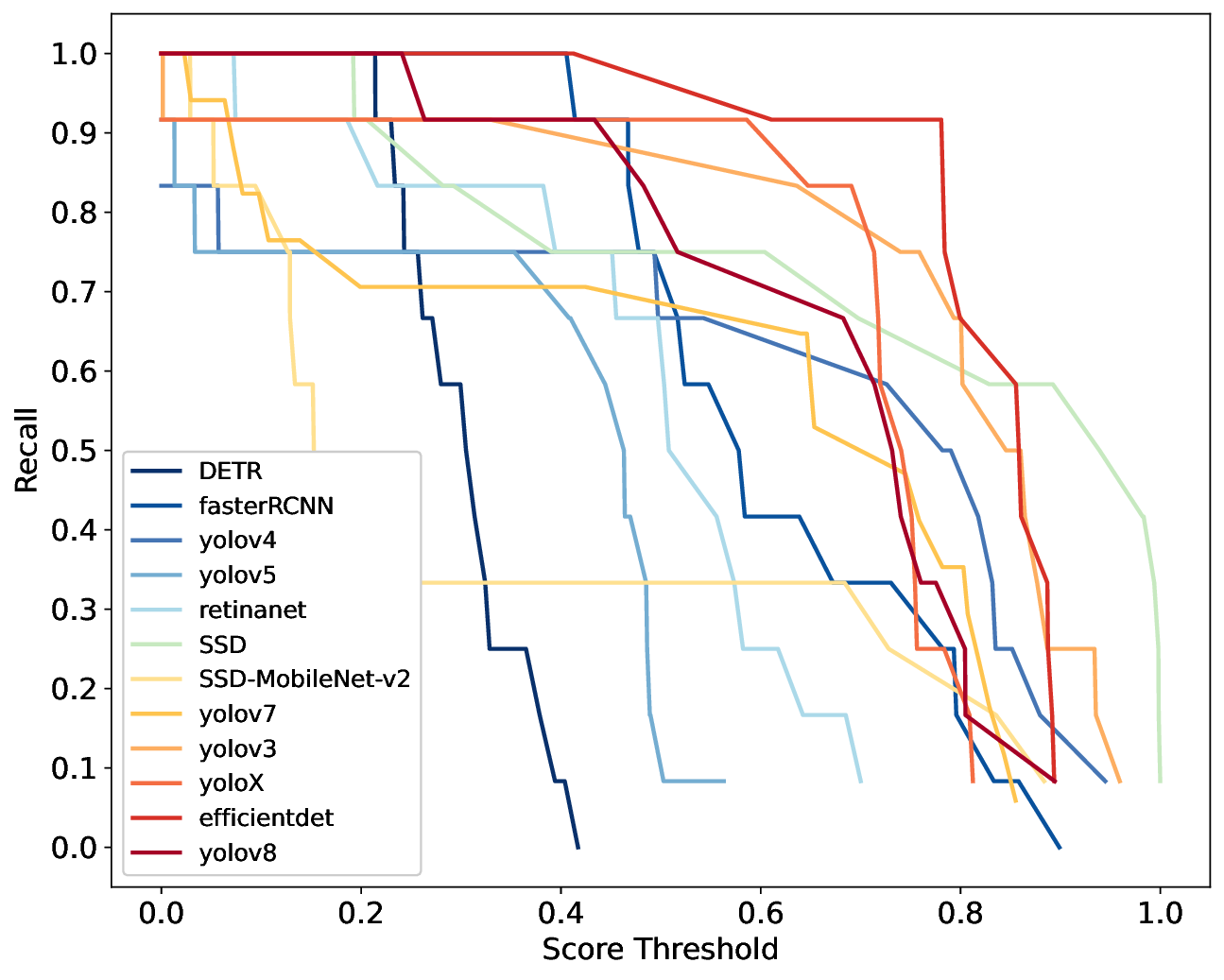}  }
\hfil
\subfloat[] {
\label{fig10b}
\includegraphics[width=0.5\linewidth,height=0.35\linewidth]{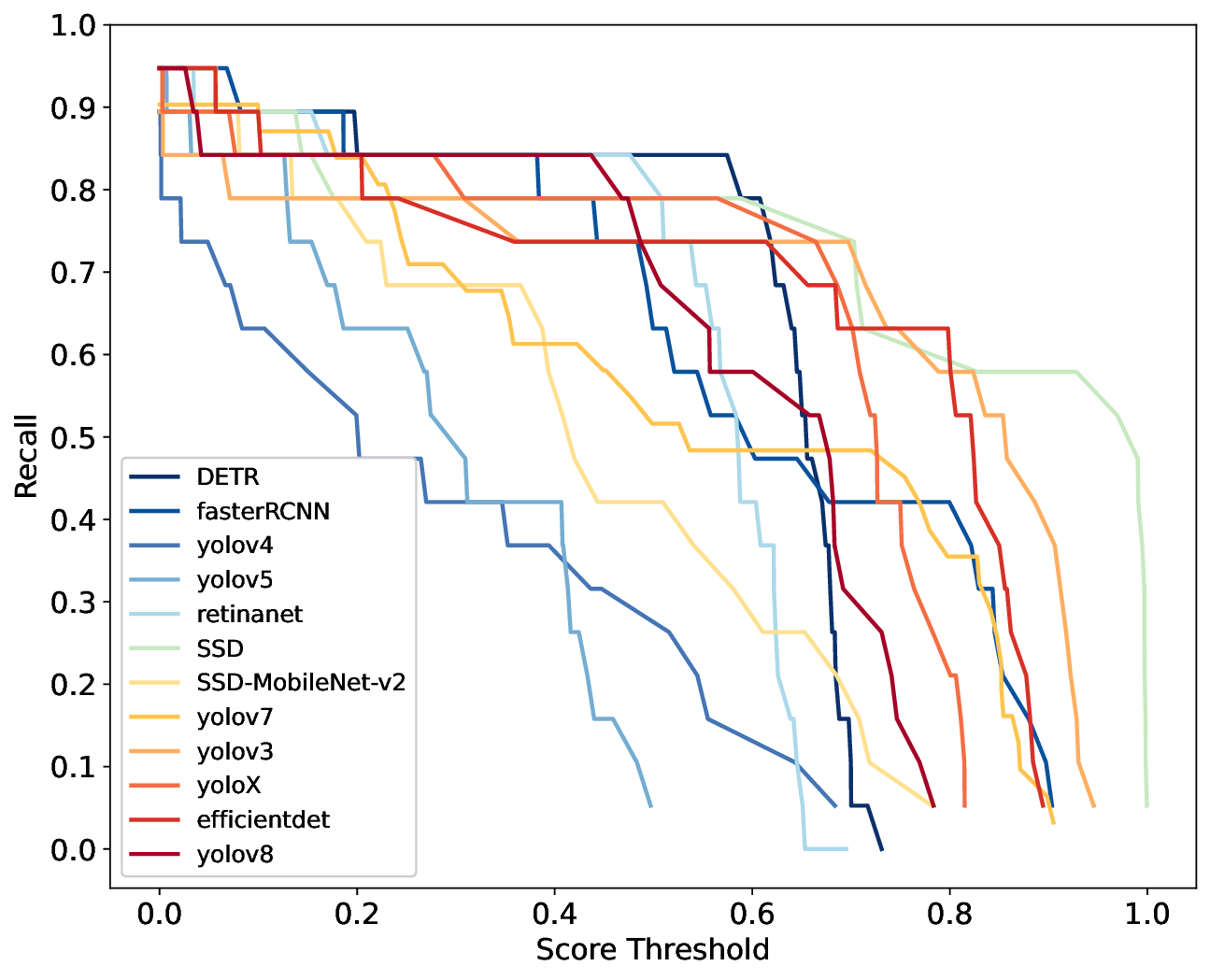}
}
\vfill
\subfloat[] {
\label{fig10c}
\includegraphics[width=0.5\linewidth,height=0.35\linewidth]{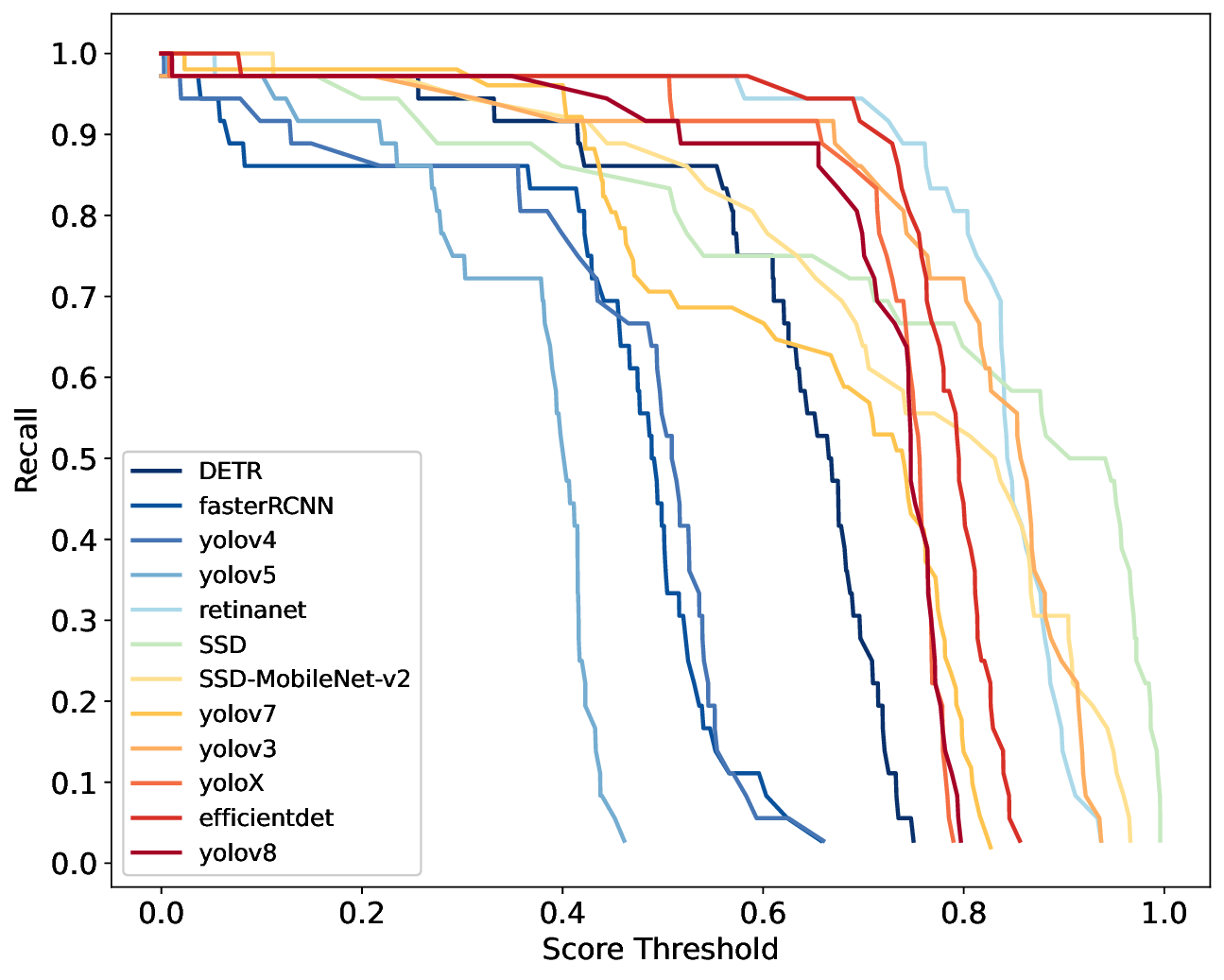}  }
\hfil
\subfloat[] {
\label{fig10d}
\includegraphics[width=0.5\linewidth,height=0.35\linewidth]{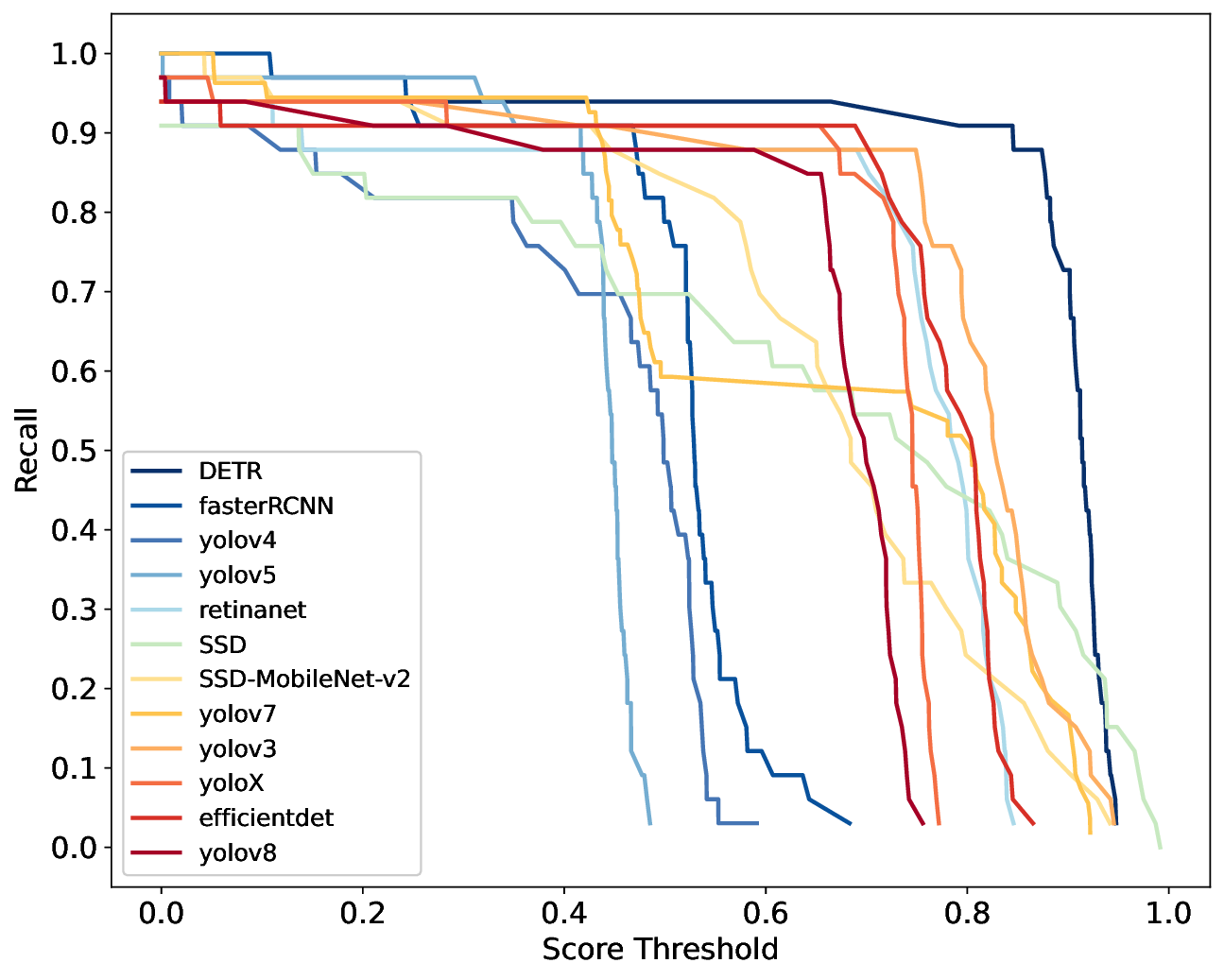}
}
\caption{Recall for detection models. (a) Left external auditory canal. (b) Right external auditory canal. (c) Left eye. (d) Right eye. Regarding the Recall metric, YOLOv8 didn't maintain its lead as it did in precision. It trailed behind both EfficientDet and YOLOX. Notably, Faster RCNN, YOLOv4, YOLOv5, SSD-MobileNet-v2, and YOLOv7 declined rapidly in recall as the threshold increases, indicating more sensitivity to threshold variations. Intriguingly, DETR, which typically underperforms, demonstrated the highest and most consistent recall for right-eye detection—the class with the most instances (312) among the four classes (312, 286, 176, 114). This hints at DETR's potential strength when applied to larger datasets. As a transformer-based architecture, DETR treats object detection as a set prediction problem. This might handle frequent or densely populated classes better than traditional architectures. Additionally, DETR captures global context from the entire image, which could be especially beneficial for detecting frequent instances scattered across the image.}
\label{fig10}
\end{minipage}
\end{figure*}

\subsection{Qualitative evaluation of Reconstruction}

A trio of experienced radiology specialists individually scrutinized both the non-standardized and standardized reconstruction outcomes, aiming to evaluate the quality of the reconstruction images. The assessment was performed with a five-tier grading scheme (1 - Subpar, 2 – Mediocre, 3 – Average, 4 – Superior, 5 – Excellent). The evaluation criteria encompassed three key elements: structural fidelity, absence of distortion, and consistency in representation across diverse viewpoints. More specifically, the experts considered whether the holistic architecture and shape of the original head were retained in the reconstructed outcomes, whether the form, contours, and features remained identifiable without distortion, and whether the reconstituted outcomes depicted a uniform interpretation of the original head, regardless of the viewpoint. To quantify the distinction between the non-standardized and standardized reconstruction outcomes, a Wilcoxon signed-rank test was employed, with the threshold of significance defined as $P <$ 0.05.

\begin{table}[tbp]
  \centering
  \setlength{\tabcolsep}{8pt}
  \caption{Qualitative evaluation of Reconstruction}
  \label{tab6}
  \begin{tabular}{ccccccc}
    \toprule
    \multirow{2}{*}{Observer} & \multirow{2}{*}{Reconstruction}& \multicolumn{5}{c}{Score}\\
    \cline{3-7}\rule{0pt}{10pt}
    & & 1 & 2 & 3 & 4 & 5\\
    \midrule
   \vspace{0.5ex}\\ [-2ex]
    \multirow{2}{*}{Observer1}  & Non-standardized & 12 & 11 & 9 & 13 & 7 \\
    & Standardized & 4 & 5 & 11 & 14 & 18\\[0.5ex] 
    \hdashline
    \vspace{0.5ex}\\ [-2ex]
    \multirow{2}{*}{Observer2}  & Non-standardized & 10 & 10 & 9 & 12 & 11 \\
    & Standardized & 2 & 4 & 10 & 15 & 21\\[0.5ex] 
    \hdashline
    \vspace{0.5ex}\\ [-2ex]
    \multirow{2}{*}{Observer3}  & Non-standardized & 12 & 11 & 6 & 10 & 13 \\
    & Standardized & 2 & 2 & 10 & 17 & 21\\[0.5ex] 
    
    \bottomrule
 \end{tabular}
\end{table}

Table \ref{tab6} shows the score distributions for both non-standardized and standardized reconstruction results. Observers 1, 2, and 3 yielded average scores with associated standard deviations for the non-standardized reconstruction results of 3.4 ± 1.0, 3.5 ± 1.0, and 3.5 ± 1.0, respectively. Meanwhile, the standardized reconstruction results correspondingly elicited scores of 4.0 ± 1.0, 4.1 ± 1.0, and 4.1 ± 1.0, each manifesting a statistically significant discrepancy at $P <$ 0.001. Among the standardized reconstruction outcomes, the quantity of cases that were assigned scores of 3 or higher, thereby being classified as clinically viable, were 43 (82.7\%), 46 (88.5\%), and 48 (92.3\%) for observers 1, 2, and 3, respectively.

\section{Discussion}

The standardized 3D reconstruction of head CT scans has profound ramifications in clinical settings. By establishing a structured, consistent, and high-resolution representation of cranial anatomy, clinicians are afforded a panoramic view of intricate structures and potential anomalies that may elude traditional imaging techniques. Therefore, our standardized 3D reconstruction holds multifaceted significance:

\begin{enumerate}
\item  \textbf{Enhanced precision in segmentation.} The standardized 3D reconstructions facilitate sharper segmentation, especially for complex structures like the brain and skull\cite{bb18}. By enabling clearer delineations, they empower clinicians to isolate specific anatomical regions with heightened accuracy, which is paramount in tasks ranging from tumor localization to post-operative assessments\cite{bb19}.

 \item  \textbf{Feature extraction \& quantitative measurements.} The reconstruction aids in spotlighting specific landmarks or features with precision. This becomes crucial when identifying and tracking the progression of specific anatomical lesions or growths. Furthermore, quantitative measurements, such as determining the volume of a tumor or assessing the length and angle of certain structures, become more feasible and precise\cite{bb20}.
 
 \item  \textbf{Alignment through 3D image registration.} Another pivotal application is in the realm of 3D medical image registration\cite{bb21}. By aligning multiple 3D images within a shared spatial domain, our reconstruction method paves the way for more holistic patient assessments, cross-referencing different imaging sessions for comprehensive insights.

 \item   \textbf{Extensibility to other anatomical regions \& modalities.} We anticipate that our approach can be extrapolated to other anatomical regions like limbs and the chest, showcasing its versatility. While some degree of model adaptation or transfer learning may be warranted for optimal outcomes, the foundational methodology remains universally applicable. Beyond CT scans, modalities like Magnetic Resonance Imaging (MRI) might benefit from our standardized reconstruction technique, thus broadening its potential impact.
 
 \end{enumerate}

Despite these advancements, our work is subject to certain limitations:

Primarily, while our standardized reconstruction gained favorable subjective outcomes in the qualitative evaluation, indicative of its utility in a clinical setting, these results are somewhat empirical. It is advisable to implement quantitative metrics to assess the loss of detail in the standardized reconstruction induced by interpolation in the reconstruction process. By treating non-standardized reconstruction performed on original CT images as the ground truth, metrics such as Mean Squared Error (MSE), Peak Signal-to-Noise Ratio (PSNR), Structural Similarity Index (SSIM), Dice Similarity Coefficient (DSC), or Jaccard Index could provide a more nuanced comparison between non-standardized and standardized reconstruction.

Secondarily, our existing data, characterized by its nuances in age, pathology, data acquisition methods, and other variables, might not fully encapsulate the intricate heterogeneity of a broader patient demographic.

This constraint in our dataset's diversity might influence the generalizability of our model. For instance, performance variations might arise when interpreting CT scans sourced from older or different imaging equipment, or when applied across diverse patient populations with unique pathologies. Such factors accentuate the need to scrutinize our model's applicability more rigorously across diverse clinical landscapes.

Given these challenges, it's paramount to enhance our dataset with a more varied collection of data. This could involve sourcing from multiple clinical settings with different scanning protocols and a wider demographic range. Incorporating diverse data types, such as MRI scans, alongside CT scans, could offer richer and more comprehensive insights. Engaging in collaborative initiatives with clinics and hospitals globally can also pave the way for a more encompassing dataset, bolstering the model's adaptability.

For those seeking immediate clinical application of our model, we recommend fine-tuning it using a subset of local data, ensuring it remains attuned to specific clinical nuances. Regular performance evaluations in the light of evolving patient demographics or changes in imaging technologies are also essential to maintain consistent accuracy.

In conclusion, to fully harness the potential of our approach across varied clinical contexts and amongst diverse patient populations, future efforts should focus on addressing these limitations and continuously refining our approach.

\section{Conclusion}\label{sec4}

In this paper, we presented a robust and efficient automated method for standardized three-dimensional reconstruction of head CT images using a deep learning-based object detection algorithm. Our solution seamlessly identifies and assesses landmarks for image reformatting, reducing inconsistencies and time demands associated with manual processes.
Through a detailed analysis of 12 object detection algorithms, we identified YOLOv8 as the most fitting choice for our task, based on reliability and efficiency.
Standardized reconstruction results further confirmed our method's clinical relevance and validity. Our innovative fusion of deep learning and radiology illuminated through this work not only holds promise for boosted diagnostic efficiency but also underscores the transformative potential of AI-driven healthcare solutions.


 
%

\bibliographystyle{IEEEtran}
\bibliography{reference.bib}

\end{document}